\documentclass[10pt,twocolumn,letterpaper]{article}

\usepackage[pagenumbers]{cvpr} % To force page numbers, e.g. for an arXiv version

\usepackage{graphicx}
\usepackage{bbding}
\usepackage{amsmath}
\usepackage{amssymb}
\usepackage[dvipsnames]{xcolor}
\usepackage{bm}
\usepackage{booktabs}
\usepackage{siunitx}
\usepackage{mhchem}

\newcommand{\yxy}[1]{\textcolor{black}{[#1]}}
\usepackage{enumitem}

\usepackage[pagebackref,breaklinks,colorlinks]{hyperref}
\usepackage[capitalize]{cleveref}
\crefname{section}{Sec.}{Secs.}
\Crefname{section}{Section}{Sections}
\Crefname{table}{Table}{Tables}
\crefname{table}{Tab.}{Tabs.}

\def\cvprPaperID{1745} % *** Enter the CVPR Paper ID here
\def\confYear{2023}

\begin{document}

%%%%%%%%% TITLE - PLEASE UPDATE
\title{Diffusion Probabilistic Model Made Slim}

\author{Xingyi Yang$^1$\quad Daquan Zhou$^2$\quad Jiashi Feng$^2$ \quad Xinchao Wang$^1$\\
National University of Singapore$^1$ \quad ByteDance Inc.$^2$\\
% Institution1 address\\
{\tt\small xyang@u.nus.edu, \{daquanzhou, jshfeng\}@bytedance.com, xinchao@nus.edu.sg}
% For a paper whose authors are all at the same institution,
% omit the following lines up until the closing ``}''.
% Additional authors and addresses can be added with ``\and'',
% just like the second author.
% To save space, use either the email address or home page, not both
% \and
% Daquan Zhou\\
% ByteDance Inc.\\
% % First line of institution2 address\\
% {\tt\small secondauthor@i2.org}
}
\maketitle
% \def\thefootnote{*}\footnotetext{Work done .}\def\thefootnote{\arabic{footnote}}
%%%%%%%%% ABSTRACT
\begin{abstract}
Despite the recent visually-pleasing results achieved,
the massive computational cost has been a long-standing flaw for
diffusion probabilistic models~(DPMs),
which, in turn, greatly limits their applications
on resource-limited platforms. 
Prior methods towards efficient DPM, however, have largely focused on accelerating the testing
yet overlooked their huge complexity and sizes.
In this paper, we make a dedicated attempt to lighten DPM while striving to preserve its favourable performance. 
We start by training a small-sized latent diffusion model~(LDM)
from scratch,
but observe a significant fidelity drop in the synthetic images.
Through a thorough assessment,
we find that DPM is intrinsically biased against high-frequency generation, and learns to recover different frequency components at different time-steps. 
These properties make compact networks unable to represent frequency dynamics with accurate high-frequency estimation. 
Towards this end, we introduce a customized design for
slim DPM, which we term as Spectral Diffusion~(\texttt{SD}),
for light-weight
image synthesis.
\texttt{SD} incorporates wavelet gating in its architecture to enable frequency dynamic feature extraction at every reverse steps, and conducts spectrum-aware distillation to promote high-frequency recovery by inverse
weighting the objective based on spectrum magni-
tudes% to \xw{xxx}
.Experimental results demonstrate that,
 \texttt{SD} achieves 8-18$\times$ computational complexity reduction as compared to the latent diffusion models on a series of conditional and unconditional image generation tasks while retaining competitive image fidelity.

% We improve the computational efficiency and image quality of diffusion models by examining its evolution in the frequency space. Our key observations are two-fold: small diffusion models have intrinsic flaw in predicting the high-frequency components of low-frequency dominant signals like natural images. On the other hand, diffusion models learn to reconstruct different frequency components at different reverse steps, thus, reduction in the steps results in spectrum outage in the final samples. For remedy those problems, we train a frequency dynamic diffusion model together with frequency-aware distillation, to  synthesizes photo-realistic images with less than $10\%$ of the original latent diffusion model and no performance compensation. For example, our model achieves xx FID on unconditional image generation on XXX dataset and $xx$ FID on class-conditioned ImageNet generation, with only XXX M model with XX s/image.

\end{abstract}

%%%%%%%%% BODY TEXT
\section{Introduction}
Diffusion Probabilistic Models~(DPMs)~\cite{ho2020denoising,song2021scorebased,song2019generative}
have recently emerged as a power tool for generative modeling,
and have demonstrated impressive results in 
image synthesis~\cite{ramesh2022hierarchical,dhariwal2021diffusion,rombach2022high},
% , hao2022magicmix}
video generation~\cite{ho2022video,ho2022imagen,zhou2022magicvideo}
and 3D editing~\cite{poole2022dreamfusion}.
Nevertheless, the gratifying results 
come with a price: 
DPMs suffer from massive model sizes.
In fact, state-of-the-art DPMs 
requires  billions of parameters,
{with hundreds or even thousands 
of inference steps per image}. 
For example, 
\emph{DALL$\cdot$ E 2}~\cite{ramesh2022hierarchical},
which is composed of 4 separate diffusion models, requires 5.5B parameters 
and 356 sampling steps in total. 
Such enormous model size, in turn,
makes DPMs extremely cumbersome to
be employed in resource-limited platforms.

\iffalse
represent an emerging topic in computer vision, providing remarkable results in the area of generative modeling, including image synthesis~\cite{ramesh2022hierarchical,dhariwal2021diffusion,rombach2022high}, video generation~\cite{ho2022video,ho2022imagen} and 3D editing~\cite{poole2022dreamfusion}. But that quality comes at a price: it is extremely cumbersome to inference a diffusion model. Current state-of-the-art diffusion model requires  billion-level parameter size, with many hundreds or thousands of model evaluations per image. For example, \emph{DALLE2}~\cite{ramesh2022hierarchical} is composed of 4 separate diffusion models, with 5.5B parameters and 356 sampling steps in total. It is hard to deploy such giant models on a modern PCs due to their enormous computational cost.
\fi
% The efficient design  has been the frontier of diffusion models.

However, existing efforts towards efficient DPMs
have  focused on model acceleration
but largely overlooked model lightening. 
For examples, the approaches of~\cite{meng2022distillation,salimans2022progressive,luhman2021knowledge,song2020denoising,bao2022analytic,liu2022pseudo,lu2022dpm}
strive for faster sampling,
while those of~\cite{ho2022cascaded,gu2022vector,vahdat2021score,rombach2022high}
rely on reducing the input size.
Admittedly, 
all of these methods give rise to 
shortened training or inference time,
yet still, the large sizes preclude them
from many real-world application scenarios.

\iffalse
To accelerate DPMs, prior efforts
have mainly concentrated
on two directions: to accelerate
sampling~\cite{meng2022distillation,salimans2022progressive,luhman2021knowledge,song2020denoising,bao2022analytic,liu2022pseudo,lu2022dpm},
or to reduce the input dimension~\cite{ho2022cascaded,gu2022vector,vahdat2021score,rombach2022high}. It is true that these methods reduce the overall running time, but the miniaturization of the large DPM remain to be a problem. A essential target of this study
is to build a extremely light-weight diffusion model with reduced parameter size and memory cost.
\fi

\begin{figure}
    \centering
    \includegraphics[width=\linewidth]{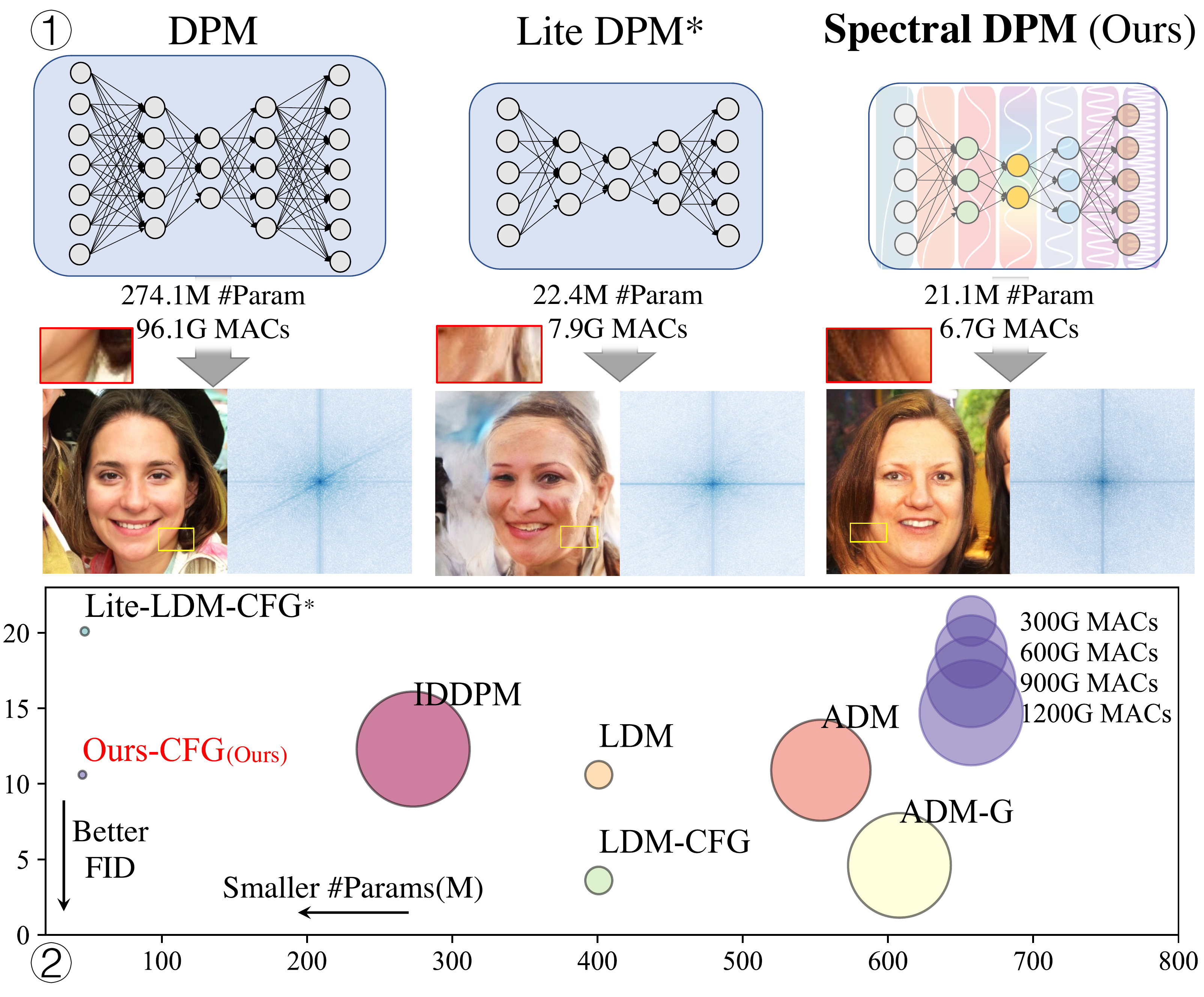}
    \vspace{-8mm}
    \caption{(1) Visualizing the frequency domain gap among generated images with the full DPM~\cite{rombach2022high}, Lite DPM and our \texttt{SD} on FFHQ~\cite{karras2019style} dataset. Lite-DPM is unable to recover fine-grained textures,
    while \texttt{SD} can produce sharp edges and realistic patterns. (2) Model size, Multiply-Add cumulation (MACs) and FID score on class-conditioned ImageNet~\cite{deng2009imagenet}. Our model achieves compelling visual quality, with minimal parameters and computational cost. $^*$ indicates our re-implemented version.}
    \label{fig:fft}
    \vspace{-5mm}
\end{figure}

In this paper, we make a dedicated attempt
towards building compact DPMs. 
To start with, we train a lite version 
of the popular latent diffusion model~(LDM)~\cite{rombach2022high}
by reducing the channel size.
We show the image generated by the original and
and lite DPM in Figure~\ref{fig:fft}.
Although the lite LDM indeed
sketches the overall structure of the faces, 
the high-frequency components, such as the skin and hair textures,
are unfortunately poorly recovered.
This phenomenon can be in fact 
revealed by the Discrete Fourier Transform~(DFT)
coefficient shown on the right column,
indicating that
the conventional design for DPMs leads to 
high-frequency deficiency when the
model is made slim.

%In addition, we plot their Discrete Fourier Transform~(DFT) coefficient on the right column, which reveals  high-frequency deficiency when training small DPM.

\iffalse
Even if our vision is clear, it is extremely difficult to train lite diffusion models from scratch. We begin by training a small version of the latent diffusion model~(LDM)~\cite{rombach2022high} by directly reducing the channel size. In Figure~\ref{fig:fft}, we plot the real and generated images with lite and official LDM. Although the lite LDM can sketch the overall structure of the faces, the high-frequency components like skin and hair textures are difficult to recover. In addition, we plot their Discrete Fourier Transform~(DFT) coefficient on the right column, which reveals  high-frequency deficiency when training small DPM.
\fi

%  We took an in-depth inspection on the generated images through the lens of frequency. It caused us to start thinking about the frequency loss in diffusion models.

We then take an in-depth inspection on the DPMs through the lens of frequency, which results in two key observations.
  (1) \textit{Frequency Evolution.} Under mild assumptions, we mathematically prove that DPMs learn different functionalities at different time-steps. 
  Specifically, we show that the optimal denoiser in fact boils down to a cascade of wiener filters~\cite{wiener1949extrapolation} with growing bandwidths. After recovering the low-frequency components, high-frequency features are added gradually in the later denoising stages. This evolution property, as a consequence, small DPMs fails to learn dynamic bandwidths with limited parameters.
(2) \textit{Frequency Bias.} DPM is biased towards dominant frequency components of the data distribution. It is most obvious when the noise amplitude is small, leading to inaccurate noise prediction at the end of the reverse process. As such, small DPMs struggle to recover the high-frequency band and image details.
%\yxy{Both property makes up the challenge to generate high-freuqncy components from small-DPM. } \xwc{linking to small models is very weak}

% Another key insights is that, 
Motivated by these observations, we propose
a novel Spectral Diffusion~(\texttt{SD}) 
model, tailored for light-weight image synthesis.
Our core idea is to 
introduce the frequency dynamics and priors into the architecture design and training objective 
of the small DPM,
so as to explicitly preserve the high-frequency details.
%which leads to a slim network for image generation without sacrificing high-frequency details. 
The proposed solution  consists of two parts, 
each accounting for one aforementioned observations.
For the frequency evolution, we propose a wavelet gating operation, which enables the network to dynamically 
adapt to the spectrum response at different time-steps. In the upsample and downsample stage, the input feature is first decomposed through wavelet transforms and the coefficients are re-weighted through a learnable gating function. It significantly lowers the parameter requirements to represent the frequency evolution in the reverse process. 

%\xwc{Transition from the last paramter to here is not smooth
{To compensate for the frequency bias for small DPMs, we distill the high-frequency knowledge from teacher DPM to a compact network.} 
This is implemented by inverse weighting the distillation loss based  on spectrum magnitudes. Specifically, 
high-frequency recovery is strengthened by over-weighting frequency bands of small magnitudes. 
Students thereby focus on the textual recovery for image generation. {By seamlessly integrating both designs, we are able to build a slim latent diffusion model, \texttt{SD}, which largely preserve the performance of LDM.} 
{Notably, \texttt{SD} by nature inherits the merits of DPMs, including superior sample diversity, training stability and tractable parameterization.} As shown in Figure~\ref{fig:fft}, our model is $8\sim 18\times$ smaller and runs $2\sim 5\times$ faster than the original LDM, while achieving  competitive image fidelity.

% \noindent\textbf{Contributions.} Our main contribution is 
The contributions of this study are threefold:
\begin{enumerate}[topsep=0pt,itemsep=0.1ex,partopsep=0.5ex,parsep=0.1ex]
    \item This study investigates the task of diffusion model slimming, which remains largely unexplored before. 
    \item We identify that the key challenge lies in its unrealistic recovery for the high-frequency components. By probing DPMs from a frequency perspective, we show that there exists a spectrum evolution over different denoising steps, and the rare frequencies cannot be accurately estimated by small models.
    % \xw{What is the difference between 1 and 2?}
    \item We propose \texttt{SD}, a slim DPM that effectively restores imagery textures by enhancing high-frequency generation performance. \texttt{SD} achieves gratifying performance on image generation tasks at a low cost.
    % \xw{Say a few more words}
\end{enumerate}

% \textbf{Motivation: }
% \begin{enumerate}
%     \item Frequency bias in Diffusion Models
%     \item Small Model exacerbate the bias. 
%     \item Temporal Dynamics of the diffusion process
% \end{enumerate}

\section{Related Work}
% \textbf{Deep Generative Models.}

\textbf{Diffusion Probabilistic Models.} DPMs~\cite{sohl2015deep,ho2020denoising} have achieved state-of-the-art results in terms of both log-likelihood estimation~\cite{song2021maximum} and sample quality~\cite{dhariwal2021diffusion}, compared to Generative adversarial Network~(GAN)-based~\cite{karras2019style,goodfellow2020generative,karras2017progressive} approaches. It has been pointed out that DPM, in its essence, is a score-based model~\cite{vincent2011connection,song2021scorebased,song2020sliced} with annealed noise scheduling~\cite{song2019generative}. The reverse process is considered as solving reverse stochastic differential equations~(SDE)~\cite{song2021scorebased}. Current best-performed DPMs are implemented as a time-conditioned UNet~\cite{ronneberger2015u,dhariwal2021diffusion,song2021scorebased} armed with self-attention~\cite{vaswani2017attention}
% zhou2021deepvit
and cross-attention~\cite{rombach2022high,huang2019ccnet}. Parameter moving average~\cite{nichol2021improved}, re-weighted objective~\cite{ho2020denoising} and advanced scheduling~\cite{nichol2021improved} significantly improves the visual quality. In this work, we focus on small diffusion designed for image generation, which has rarely been studied before. 

\textbf{Efficient Diffusion.} The efficient diffusion model for low-resource inferences has recently become a popular research topic.
One approach 
is through reducing 
the sampling steps, 
which is either done 
by distilling multiple 
steps into a single step~\cite{meng2022distillation,salimans2022progressive,luhman2021knowledge}, 
or shortening the reverse steps 
while maintaining the image fidelity~\cite{song2020denoising,bao2022analytic,liu2022pseudo,lu2022dpm}. 
Another possible solution 
explores 
the idea of 
diffusing in a lower dimensional space
, and then scaling it up, 
with a cascade structure~\cite{ho2022cascaded} or in the latent space~\cite{vahdat2021score, rombach2022high}. In distinction from them, we build an efficient diffusion model 
using light-weight architecture and knowledge distillation.

\textbf{Frequency Analysis for Generative Model.} 
Neural networks tend
to fit low-frequency signals first 
and shift to the high-frequency components, which is referred to as \emph{frequency principle} of deep neural network~\cite{xu2019frequency,xu2019training,basri2020frequency}. 
The frequency bias 
is also observed 
when training deep generative models like GANs~\cite{frank2020leveraging,chen2021ssd,khayatkhoei2022spatial,schwarz2021frequency}, where the generator 
struggles to build up 
natural high-frequency details.

In this paper, we examine the frequency behavior of DPMs.
Taking advantage of its frequency properties, our \texttt{SD} achieves realistic image generation at a low cost. 
\section{Background}
\subsection{Denoising Diffusion Probabilistic Models}

Diffusion model reverses a progressive noise process based on latent variables. Given data $\mathbf{x}_0 \sim q(\mathbf{x}_0)$ sampled from the real distribution, we consider perturbing data with Gaussian noise with zero mean and $\beta_t$ variance for $T$ steps
\begin{align}
    q(\mathbf{x}_t|\mathbf{x}_{t-1})=\mathcal{N}(\mathbf{x}_{t};\sqrt{1-\beta_{t}}\mathbf{x}_{t-1}, \beta_{t}\mathbf{I})
\end{align}
where $t\in [1, T]$ and $0<\beta_{1:T}<1$ denote the noise scale scheduling. At the end of day, $\mathbf{x}_T \to \mathcal{N}(0, \mathbf{I})$ converge to a Gaussian white noise. Although sampling from noise-perturbed distribution $q(\mathbf{x}_t)=\int q(\mathbf{x}_{1:t}|\mathbf{x}_0) d \mathbf{x}_{1:t-1}$ requires a tedious numerical integration over steps, the choice of Gaussian noise provides a close-form solution to generate arbitrary time-step $\mathbf{x}_t$ through
\begin{align}
    \mathbf{x}_t = \sqrt{\bar{\alpha}} \mathbf{x}_0 + \sqrt{1-\bar{\alpha}} \bm{\epsilon}, \quad \text{where} \quad \epsilon \sim \mathcal{N}(0, \mathbf{I})
\end{align}
where $\alpha_t = 1-\beta_t$ and $\bar{\alpha}_t = \prod_{s=1}^t \alpha_s$. A variational Markov chain in the reverse process is parameterized as a time-conditioned denoising neural network $\mathbf{s}(\mathbf{x}, t;\bm \theta)$ with $p_{\bm \theta}(\mathbf{x}_{t-1}|\mathbf{x}_t)=\mathcal{N}(\mathbf{x}_{t-1}; \frac{1}{\sqrt{1-\beta_t}}(\mathbf{x}_t+\beta_t \mathbf{s}(\mathbf{x}_t, t;\bm \theta)), \beta_t \mathbf{I})$. The denoiser is trained to minimize a re-weighted evidence lower bound~(ELBO) that fits the noise
\begin{align}
    \mathcal{L}_{\text{DDPM}} &= \mathbb{E}_{t,\mathbf{x}_0, \bm\epsilon} \Big[||\bm\epsilon - \mathbf{s}(\mathbf{x}_t, t;\bm \theta) ||_2^2\Big]\label{eq:loss_ddpm}\\& = \mathbb{E}_{t,\mathbf{x}_0, \bm\epsilon} \Big[||\nabla_{\mathbf{x}_t} \log p(\mathbf{x}_t|\mathbf{x}_0) - \mathbf{s}(\mathbf{x}_t, t;\bm \theta) ||_2^2\Big]
\end{align}
where the $\nabla_{\mathbf{x}_t} \log p(\mathbf{x}_t|\mathbf{x}_0)$ are also called the score function~\cite{song2019generative}. Thus, the denoiser equivalently learns to recover the derivative that maximize the data log-likelihood~\cite{hyvarinen2005estimation,vincent2011connection}.
With a trained $\mathbf{s}(\mathbf{x}, t;\bm \theta^*)\approx\nabla_{\mathbf{x}_t} \log p(\mathbf{x}_t|\mathbf{x}_0)$, we  generate the data by reversing the Markov chain
\begin{align}
    \mathbf{x}_{t-1}\leftarrow \frac{1}{\sqrt{1-\beta_t}}(\mathbf{x}_t+\beta_t \mathbf{s}(\mathbf{x}_t, t;\bm \theta)) + \sqrt{\beta_t} \bm \epsilon_t
\end{align}
The reverse process could be understood as going along $\nabla_{\mathbf{x}_t} \log p(\mathbf{x}_t|\mathbf{x}_0)$ from $\mathbf{x}_T$ to maximize the data likelihood.

\subsection{Frequency Domain Representation of Images}
Frequency domain analysis decomposes a image according to a sets of basis functions. We focus on two discrete transformations: \emph{Fourier} and \emph{Wavelet} Transform. 

Given a $H\times W$ input signal\footnote{For simplicity, we only introduce the formulation for gray-image, while it is extendable to multi-channel inputs.} $\mathbf{x} \in \mathbb{R}^{H\times W}$, Discrete Fourier Transform~(DFT) $\mathcal{F}$ projects it onto a collections of sine and cosine waves of different frequencies and phases
{\small\begin{align*}
   \mathcal{X}(u,v) = \mathcal{F}[\mathbf{x}] = \sum_{x=1}^H\sum_{y=1}^W \mathbf{x}(x,y) e^{-j 2\pi (\frac{u}{H}x + \frac{v}{W}y)}
\end{align*}}
$\mathbf{x}(x,y)$ is
the pixel value at $(x,y)$; $\mathcal{X}(u, v)$ represents complex value at frequency $(u, v)$; $e$ and $j$ are Euler’s number and the imaginary unit.

On the other hand, Discrete Wavelet Transform~(DWT) projects it onto multi-resolution wavelets functions. In a singlescale case, $\mathbf{x}$ is decomposed into 4 wavelet coefficients $\mathbf{x}_\textsf{LL}, \mathbf{x}_\textsf{LH}, \mathbf{x}_\textsf{HL}, \mathbf{x}_\textsf{HH}=\textsf{DWT}(\mathbf{X})$ with halving the scale, where $\mathbf{x}_{\{\textsf{LL},\textsf{LH},\textsf{HL}, \textsf{HH}\}} \in \mathbb{R}^{\frac{H}{2}\times \frac{W}{2}}$. $\mathbf{x}_\textsf{LL}$ stands for low-frequency component and $\mathbf{x}_{\{\textsf{LH},\textsf{HL}, \textsf{HH}\}}$ are high-frequency components that contains the textural details. The coefficients could then be inverted and up-sampled back to the original input $\mathbf{x}=\textsf{IDWT}(\mathbf{x}_\textsf{LL}, \mathbf{x}_\textsf{LH}, \mathbf{x}_\textsf{HL}, \mathbf{x}_\textsf{HH})$. 
\section{Frequency Perspective for Diffusion}
In general signal processing, denoising is often performed in frequency space.
Similar to Figure~\ref{fig:fft}, Table~\ref{tab:freq_error} compares Low-freq and High-freq error\footnote{The error computed as the $\mathbb{E}_{f}[\mathbb{E}[|\mathcal{F}_{real}|]-\mathbb{E}[|\mathcal{F}_{gen}|]]$ over 300 real and generated samples, with the low-high cut-off frequency of 28Hz.} for different DPMs on FFHQ dataset. Lite-LDM performs poorly due to its lack of high-frequency generation.
\begin{table}[h]
\footnotesize
\vspace{-2mm}
    \centering
    \begin{tabular}{l|l|c|l|l}
     \hline
        Method & \#Param& FID$\downarrow$ & Low-freq Error$\downarrow$ & High-freq Error$\downarrow$ \\
        \hline
        LDM &274.1M& 5.0 &0.11 & 0.75 \\
        Lite-LDM &22.4M& 17.3 & 0.28(\textcolor{Red}{+0.17}) & 3.35(\textcolor{Red}{+2.17})\\
         \hline
    \end{tabular}
    \vspace{-2mm}
    \caption{Low-freq and High-freq error for different model size. }
    \label{tab:freq_error}
    \vspace{-2mm}
\end{table}

\begin{figure}[t]
    \centering
    \includegraphics[width=\linewidth]{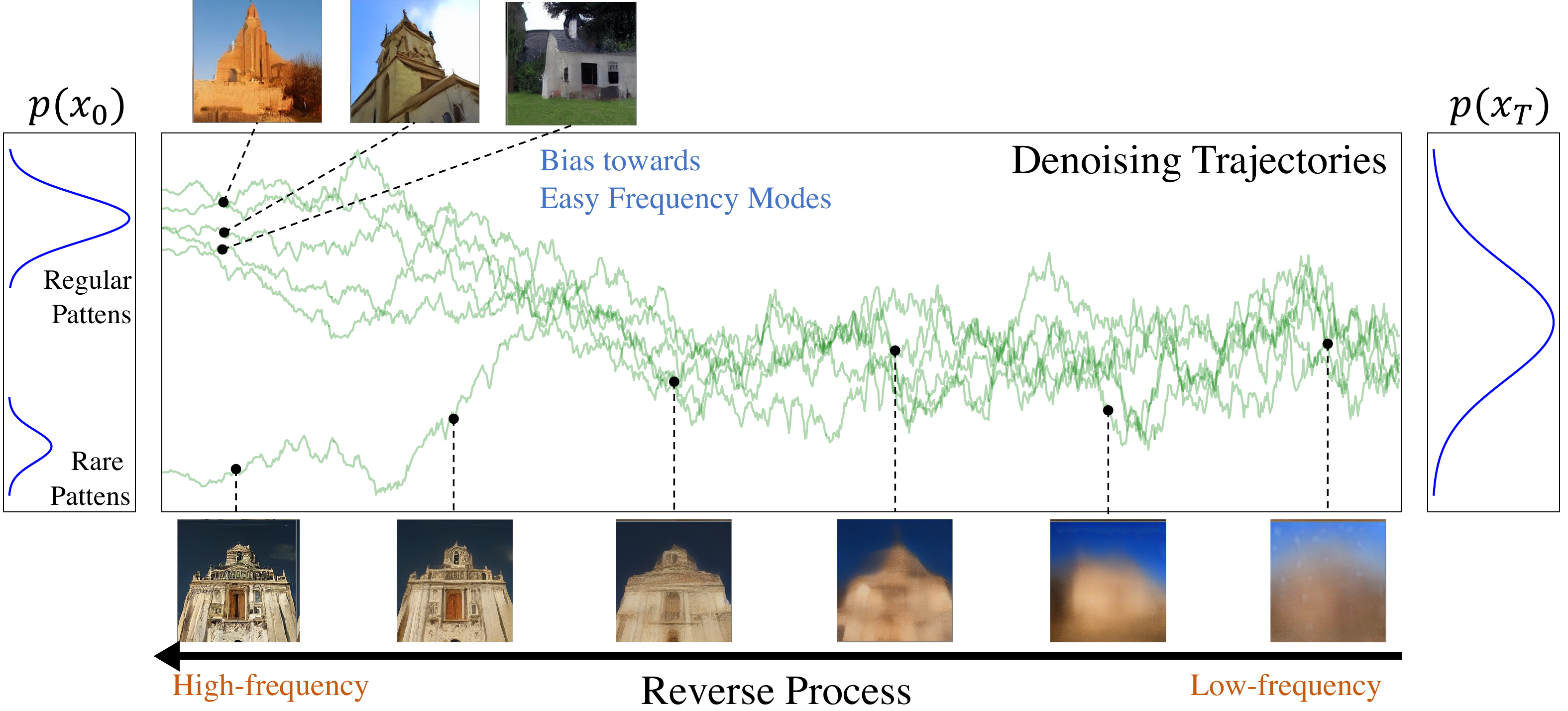}
    \vspace{-5mm}
    \caption{\textbf{Illustration of the Frequency Evolution and Bias for Diffusion Models.} In the reverse process, the optimal filters recover low-frequency components first and add on the details at the end. The predicted score functions may be incorrect for rare patterns, thus failing to recover complex and fine-grained textures.}
    \label{fig:trajectory}
    \vspace{-5mm}
\end{figure}

Thus, we examine DPM's behavior in the frequency domain.
As illustrated in Figure~\ref{fig:trajectory}, we make two findings: (1) \emph{Frequency Evolution.} Diffusion model learns to recover the low-frequency components at first, and gradually adds in photo-realistic and high-frequency details. (2) \emph{Frequency Bias.} Diffusion model makes biased recovery for the minority frequency band.
% \begin{enumerate}
%     \item \textit{What is recovered?} 
%     \item \textit{What cannot be recovered?} 
% \end{enumerate}
% \subsection{Small DPM Fails to Recover High-frequency}
% Our discussion starts by training a Lite-version LDM~\cite{rombach2022high} by reducing the channel number on FFHQ dataset. We compare its performance with the official LDM. The Lite-LDM only achieves FID score of 17.3, compared to the original LDM with 5.0 FID. We thus compared take 

\subsection{Spectrum Evolution over Time}
\label{sec:freq_evo}
% A diffusion process~\cite{ho2020denoising} first maps data to noise by gradually perturbing the input data with additive Gaussian noise, and then recover original data through a iterative denoising reverse process. Natural images are always dominant by low-frequency components, and Gaussian noise has uniform power spectrum across all frequency components. Those structural properties drive us to inspect the diffusion process in frequency domain. Our key observation is that, diffusion model learns to recover the low-frequency components when $t$ is large, and gradually add in photo-realistic and high-frequency details  when $t$ is small. 

% Let's start with an additive white noise model $\mathbf{x}' = \mathbf{x} + \bm \epsilon$,
% where $\mathbf{x}$ is the clean signal and $\bm \epsilon \sim \mathcal{N}(0,\sigma^2 \mathbf{I})$ is a Gaussian white noise. A optimal denoiser estimates the recovered signal that best recover $\mathbf{x}$. We take a different notation to recover noise $\bm\epsilon \approx \mathbf{s}(\mathbf{x}';\bm \theta)$ to align with DPM objective
% \begin{align}
%     \bm \theta^* = \underset{\bm \theta, \mathbf{x}}{\arg \min} \mathbb{E}_{\bm \theta}[ ||\mathbf{s}(\mathbf{x} + \bm \epsilon;\bm \theta) - \bm \epsilon||_2^2]
% \end{align}
% Note that both predicting the signal or the noise does not make any difference in the additive model, since the reconstructed image is $\mathbf{x}'-\mathbf{s}(\mathbf{x}';\bm \theta)$. 
DPM optimizes a time-conditioned network to fit noise at multiple scales, which gives rise to a denoising trajectory over time-steps. We take a close look at this trajectory through the lens of frequency. When assuming the network is a linear filter, we give the optimal filter in terms of its spectrum response at every timestep. This filter is commonly known as \textbf{Wiener filter}~\cite{wiener1949extrapolation}.

% \noindent\fbox{
% \centering
% \noindent\begin{minipage}{0.95\linewidth}
\noindent\textbf{Proposition 1.} \textit{Assume $\mathbf{x}_0$ is a wide-sense stationary signal and $\bm \epsilon$ is white noise of variance $\sigma^2=1$. For $\mathbf{x}_t = \sqrt{\bar{\alpha}} \mathbf{x}_0 + \sqrt{1-\bar{\alpha}} \bm{\epsilon}$, the optimal linear denoising filter $h_t$ at time $t$ that minimize $J_t=\|h_t \ast \mathbf{x}_t - \bm \epsilon\|^2$ has a closed-form solution} 
\begin{align}
    \mathcal{H}_{t}^*(f) = \frac{1}{\bar{\alpha}|\mathcal{X}_0(f)|^2 + 1-\bar{\alpha}}
\end{align}
\textit{where $|\mathcal{X}_0(f)|^2$ is the power spectrum of $\mathbf{x}_0$
and $\mathcal{H}^*
_t (f)$ is the frequency response of $h_t^*$. }

Although the linear assumption poses a strong restriction on the model architecture, we believe it provides valuable insights into what has been done in the reverse process. 

\begin{figure}
    \centering
    \includegraphics[width=\linewidth]{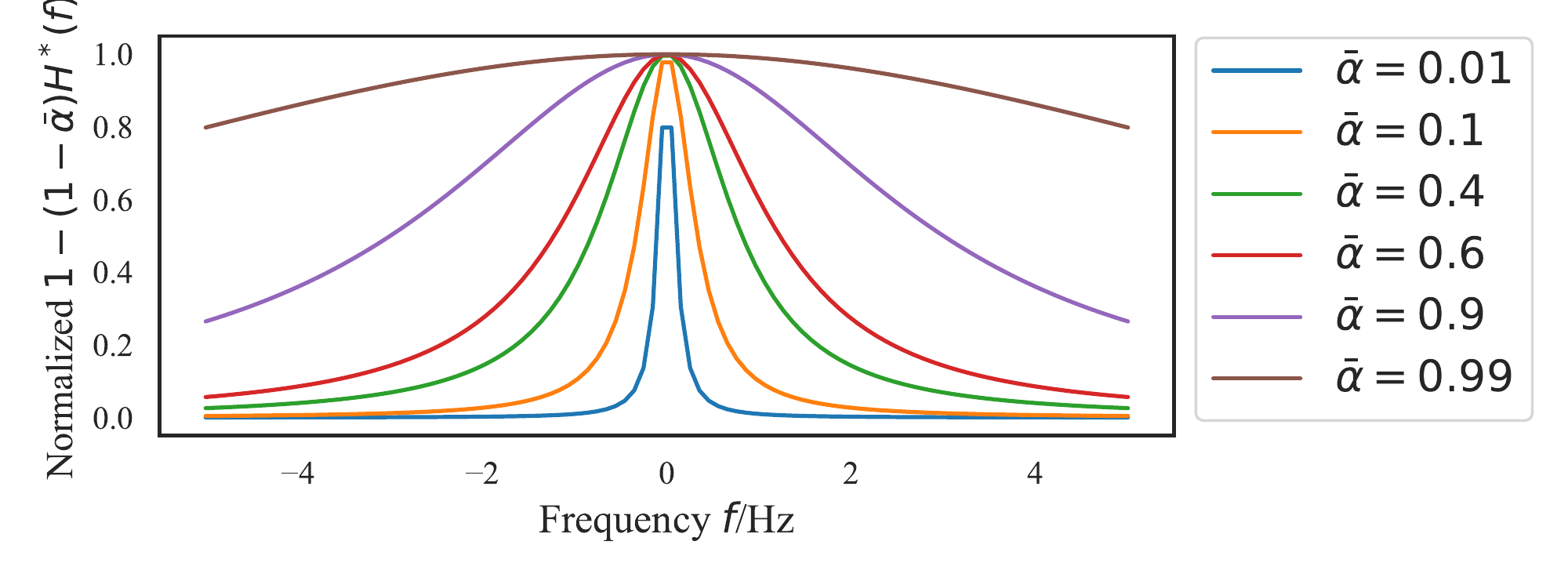}
    \vspace{-8mm}
    \caption{$1-(1-\bar{\alpha})|H^*(f)|^2$ of the optimal linear denoising filter with different $\bar{\alpha}$.}
    \label{fig:wiener_filter}
    \vspace{-6mm}
\end{figure}

\noindent\textbf{DPM goes from structure to details.} In this study, we make a widely accepted assumption about the power spectra of natural images  $\mathbb{E}[|X_0(f)|^2] = A_s(\theta)/f^{\alpha_S(\theta)}$that follows a power law~\cite{van1996modelling,burton1987color,field1987relations,tolhurst1992amplitude}. $A_s(\theta)$ is called an amplitude scaling factor and $\alpha_S(\theta)$ is the frequency exponent. If we set $A_s(\theta)=1$ and $\alpha_S(\theta)=2$, the frequency response of the signal reconstruction filter $1-\sqrt{1-\bar{\alpha}}h$ is in Figure~\ref{fig:wiener_filter}.

In the reverse process, $t$ goes from $T\to 0$, and $\bar{\alpha}$ increases from $0\to 1$. Therefore, DPM displays a spectrum-varying behavior over time. In the beginning, we have a narrow-banded filter ($\bar{\alpha}= 0.1$ and $\bar{\alpha}=0.01$) that only restores the low-frequency components that control the rough structures. $t$ goes down and $\bar{\alpha}$ gradually increases, with more details and high-frequency components restored in the images, like the human hairs, wrinkles, and pores.

We plot the denoised predictions $\hat{\mathbf{x}}_0$ at different steps using pre-trained LDM~\cite{rombach2022high} in Figure~\ref{fig:trajectory}, which shows that DPM generates low-frequency first and transits into high-frequency. The same empirical observation that DPM goes from rough to details has been shown in~\cite{ho2020denoising,ma2022pds,choi2022perception,rombach2022high}, while we are the first to give its numerical solutions.

\begin{figure}
    \centering
    \includegraphics[width=\linewidth]{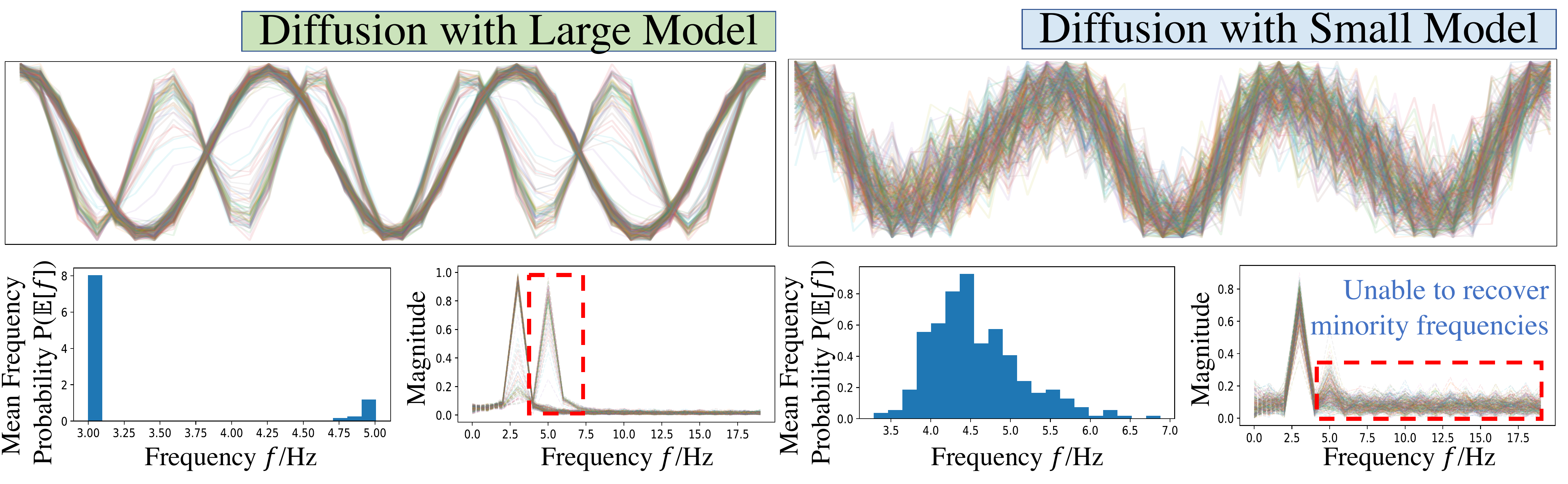}
    \caption{Toy example for 1D signal fitting. Small DPM is unable to recover minority frequency components.}
    \label{fig:freq_bis}
    \vspace{-4mm}
\end{figure}
\subsection{Frequency Bias in Diffusion Model}\label{sec:freq_bias}
Another challenge in diffusion-based model is the inaccurate denoising estimation in low-density regions~\cite{song2019generative}. It results from the expectation over $p(\mathbf{x}_0)$ in the loss function
{\small\begin{align}
    \mathcal{L}_{\text{DDPM}} = \int p(\mathbf{x}_0)\mathbb{E}_{t, \bm\epsilon} \Big[||\bm\epsilon - \mathbf{s}(\mathbf{x}_t, t;\bm \theta) ||_2^2\Big] \text{d}\mathbf{x}_0
\end{align}}
Since the denoising objective is weighted by $p(\mathbf{x}_0)$, the trained diffusion will be biased towards the high-density region, while ignoring the long-tail patterns. 

For image generation tasks, one long-tail pattern is the frequency bias. While the low-frequency images are dominant with large $p(\mathbf{x}_0)$, very few samples contain high-frequency components. Training small DPMs on the biased data distribution makes it difficult to generate samples with complex textures and realistic high-frequency patterns.

% \noindent\fbox{
% \centering
% \noindent\begin{minipage}{0.95\linewidth}
\noindent\textbf{Example 1.} We fit a toy diffusion model to 1D functions $f(x)=cos(\alpha 2\pi x)$, where $P(\alpha=3)=0.2$ and $P(\alpha=5)=0.8$. We adopt a two-layer feed-forward neural network, with 1000 denoising steps and hidden units $M=\{64, 1024\}$. More details in in Supplementary.

We plot the 300 generated signals in Figure~\ref{fig:freq_bis}~(Top), their DFT magnitudes in (Button Right), and the mean frequency histogram in (Button Left). Small model~($M=64$) faces difficulty recovering the minority frequencies other than $\alpha=3$, while large model~($M=1024$) achieves smooth denoised results over all freq bands, especially when $\alpha=5$.
% \end{minipage}
% }

It provides concrete evidence that small DPMs have intrinsic defects in recovering the high frequencies.

\section{Spectral Diffusion Model}
As explained above, our goal is to slim down the DPMs by introducing the frequency dynamics and priors into the architecture design and training objectives. Taking the LDM~\cite{rombach2022high} as our baseline, we design a wavelet-gating module to enable time-dynamic inference for the network with a limited model size. A spectrum-aware distillation is applied to enhance the high-frequency generation performance. Both modifications allow us to achieve photo-realistic image generation with minimal model size and computational effort.

% \subsection{Baseline for }

\subsection{Dynamic Wavelet Gating}
As depicted in Section~\ref{sec:freq_evo}, the reverse process requires a cascade of filters with dynamic frequency response. Vanilla UNet~\cite{ronneberger2015u}, while being effective in reconstructing image details, is incapable to incorporate dynamic spectrum into a single set of parameters. As a result, the small-size DPM is incapable to compensate for the changing bandwidth. 

In response to such frequency evolution, we propose to insert the \textbf{Wavelet Gating}~(WG) module into the network to automatically adapt it to varing frequency response. WG decomposes the feature map into wavelet bands and selectively attends to the proper frequency at different reverse steps, which is uniquely tailored for the diffusion model.

% We choose \textsf{DWT} since \textsf{DWT} is generally faster to compute than other frequency operator like Discrete Cosine Transform~(DCT) and Fast Fourier Transform~(FFT). 

\begin{figure}
    \centering
    \includegraphics[width=\linewidth]{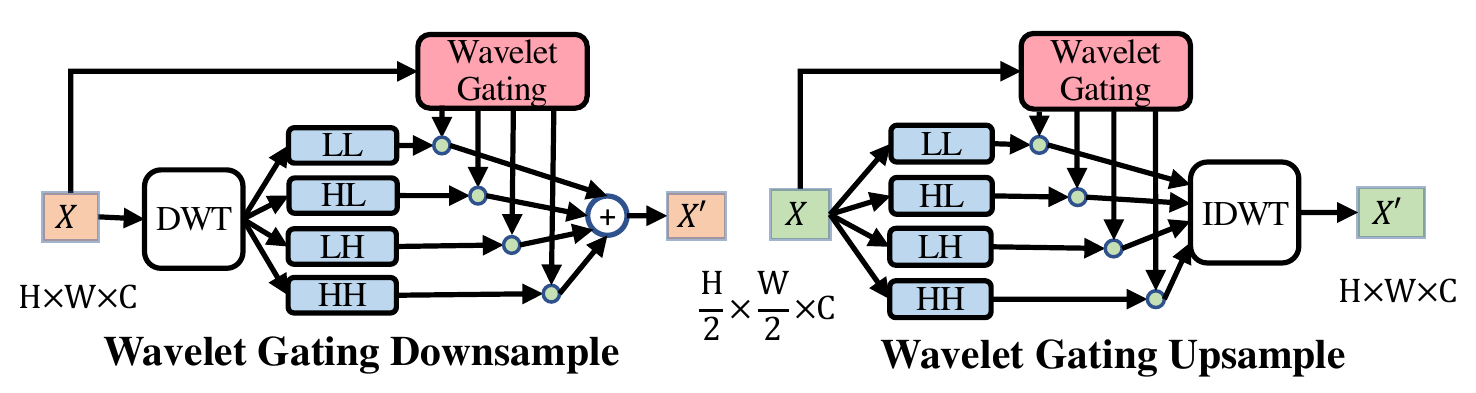}
    \vspace{-6mm}
    \caption{\texttt{WG-Down} and \texttt{WG-Up} with wavelet gating.}
    \vspace{-4mm}
    \label{fig:wg}
\end{figure}

\noindent\textbf{Gating over the Wavelet Coefficients.} We replace all down-sample and up-sample in UNet with \textsf{DWT} and \textsf{IDWT}~\cite{fu2021dw,yang2022wavegan}, and pose a soft gating operation on wavelet coefficients to facilitate step-adaptive image denoising. We call them \texttt{WG-Down} and \texttt{WG-Up}, as shown in Figure~\ref{fig:wg}.

Following the channel attention operation~\cite{woo2018cbam,hu2018squeeze,qin2021fcanet}, information from input feature $\mathbf{X}$ is aggregated to produce a soft gating mask
\begin{align}
    g_{\{\textsf{LL},\textsf{LH},\textsf{HL}, \textsf{HH}\}} &= \text{Sigmoid}(\text{FFN}(\text{Avgpool}(\mathbf{X})))
\end{align}
where $g_{i}$ is the gating score of each wavelet band; $\text{FFN}$ is a 2 layer where feed-forward network and $\text{Avgpool}$ stands for the average pooling. The coefficients are then gated with $g_{i}$ to produce the output $\mathbf{X}'$. 

In the \texttt{WG-Down}, we apply WG after the \textsf{DWT} operation to fuse the sub-band coefficients with weighted summation $ \mathbf{X}' = \sum_{i\in\{\textsf{LL},\textsf{LH},\textsf{HL}, \textsf{HH}\}} g_{i} \odot \mathbf{X}_{i}$, where $\odot$ is the element-wise multiplication. In the  \texttt{WG-Up}, the input feature is splitted into 4 chunks as the wavelet coefficients. Then, WG is carried out to re-weight each sub-band before $\mathbf{X}'= \textsf{IDWT}(g_\textsf{LL} \odot\mathbf{X}_\textsf{LL}, g_\textsf{LH} \odot\mathbf{X}_\textsf{LH}, g_\textsf{HL} \odot\mathbf{X}_\textsf{HL}, g_\textsf{HH} \odot\mathbf{X}_\textsf{HH})$. In this paper, we apply Haar wavelet by default.

\subsection{Spectrum-Aware Knowledge Distillation}
Diffusion model has difficulty in modelling the high-frequency components~(in Section~\ref{sec:freq_bias}), especially for efficient requirements. In combat with spectrum deficiency in image generation, we distill the prediction of a large pre-trained teacher model to a compact WG-Unet student. Beyond output matching with a L2 loss, a \textbf{Spectrum-Aware Distillation} is applied to guide the student to synthesize naturalistic image details. Our intuition is to re-weight the distillation loss according to the spectrum magnitude. For components with low magnitudes, such as high-frequency bands, we increase the error penalty; while the weight for the low-frequency elements is reduced. 

Suppose a teacher diffusion model $\bm s_T(\cdot; \bm \theta_T)$, we would like to distill a student $\bm s_T(\cdot; \bm \theta_T)$ by mimicking the outputs and features
% ~\cite{hinton2015distilling,yang2022factorizing}
. At time-step $t$, the perturbed image $\textbf{x}_t$ is fed into both networks to produce the outputs and features. A L2 loss~\cite{romero2014fitnets,liu2019structured} is use to quantify their spatial distance 
\begin{align}
    \mathcal{L}_{\text{spatial}} = \sum_{i} \|\mathbf{X}^{(i)}_T- \mathbf{X}^{(i)}_S \|_2^2
\end{align}
where $\mathbf{X}^{(i)}_{T}$ and $\mathbf{X}^{(i)}_{S}$ stand for the pair of teacher/student's output features or outputs of the same scale. A single 1$\times$1 \textsc{Conv} layer is used to align the dimensions between a prediction pair. 
% $\lambda_s$ is the weight to balance the style loss and the L2 loss. 
% Note that, we matching the output channels of each feature pairs Both losses contribute to spatial reconstruction of the teacher outputs with the student response.

In addition to the spatial distillation, inspired by the imbalanced learning~\cite{lin2017focal,cao2019learning,jiang2021focal} and long-tail learning~\cite{zhang2021deep,Kang2020Decoupling}, we design a distillation loss to encourage the model for minority frequency recovery. Given a pair of model predictions and the clean image $\mathbf{x}_0$, we first interpolate $\mathbf{x}_0$ to the same size of the feature map, then take their 2D DFT
{\small
\begin{align}
    \mathcal{X}^{(i)}_{T}=\mathcal{F}[\mathbf{X}^{(i)}_{T}],\mathcal{X}^{(i)}_{S}=\mathcal{F}[\mathbf{X}^{(i)}_{S}], \mathcal{X}^{(i)} =\mathcal{F}[\text{Resize}(\mathbf{x}_0)]
\end{align}}
The $\mathcal{X}_{0}$ is then applied to modulate the difference between $\mathcal{X}^{(i)}_{T}$ and $\mathcal{X}^{(j)}_{S}$
\begin{align}
    \mathcal{L}_{\text{freq}} = \frac{}{} \sum_{i} \omega_{i} \|\mathcal{X}^{(i)}_{T}- \mathcal{X}^{(j)}_{S} \|_2^2, \text{where } \omega = |\mathcal{X}^{(i)}|^\alpha
\end{align}
with a scaling factor $\alpha < 0$ ($\alpha= -1$ in our experiment), $\mathcal{L}_{\text{freq}}$ pushes the student towards learning the minority frequencies yet down-weights the majority components. Together with the DDPM objective in Eq.~\ref{eq:loss_ddpm}, our training objective becomes $\mathcal{L}=\mathcal{L}_{\text{DDPM}}+\lambda_s \mathcal{L}_{\text{spatial}} + \lambda_f \mathcal{L}_{\text{freq}}$ with weighting factors $\lambda_s=0.1$ and $\lambda_f =0.1$.

Note that our method aims to learn accurate score prediction at each denoising step, which is orthogonal to existing distillation on  sampling step reduction~\cite{salimans2022progressive,meng2022distillation}.
\section{Experiments}
This section demonstrates the ability of our approach \texttt{SD} on high-resolution image synthesis~(Section \ref{sec:uncondition}) with limited computation, and validates the significance of each proposed module via ablation study in Section~\ref{sec:ablation}.
\begin{table*}[]
\begin{minipage}{0.71\textwidth}
\renewcommand{\arraystretch}{1.}
\footnotesize
{\setlength{\tabcolsep}{1pt}
    \centering
    \begin{tabular}{l|l|l|c}
    \hline
    \multicolumn{4}{c}{\textbf{FFHQ $256\times 256$}}\\
    \hline
     Model &\#Param & MACs & FID$\downarrow$  \\
      \hline
    %  BigGAN~\cite{brock2018large} & GAN & 55.88M & 112.79G &12.4 \\
    %  StyleGAN~\cite{karras2019style} & & & 4.4 & \\
    %  StyleGAN2~\cite{karras2020analyzing}& GAN & 72.03M & 143.15G & 4.5  \\
    %   VQGAN~\cite{esser2021taming} & GAN+AR & & & 9.6 \\
    
     DDPM~\cite{ho2020denoising}  & 113.7M &248.7G & 8.4\\
    P2~\cite{choi2022perception}  & 113.7M &248.7G & 7.0\\
    LDM~\cite{rombach2022high}  & 274.1M & 96.1G & 5.0\\
     \hline
    %  GAN-Slim~\cite{wang2020gan} & GAN & & 5.0G & 12.4 \\
    %  CA-Comp~\cite{liu2021content} & GAN & & 4.1G & 7.9 \\
     Lite-LDM & 22.4M({\tiny\textcolor{Green}{$ 12.2\times$}}) & 7.9G({\tiny\textcolor{Green}{$12.2\times $}}) & 17.3({\tiny\textcolor{red}{$-12.3$}})\\
     Ours & 21.1M({\tiny\textcolor{Green}{$13.0\times$}})& 6.7G({\tiny\textcolor{Green}{$ 14.3\times$}}) & 10.5({\tiny\textcolor{Orange}{$-5.5$}})\\
     \hline
    \end{tabular}
     \hspace{0em}
    \begin{tabular}{l|l|l|c}
    \hline
    \multicolumn{4}{c}{\textbf{CelebA-HQ $256\times 256$}}\\
    \hline
     Model &\#Param & MACs & FID$\downarrow$  \\
      \hline
    %   NVAE~\cite{vahdat2020nvae} & VAE & 330.79M& &  29.76\\
    %   PGGAN~\cite{karras2018progressive} & GAN & 23.05M & 28.17G & 8.0\\
    %  VQGAN~\cite{esser2021taming} & GAN+AR & & & 10.7 \\
    %  StyleGAN~\cite{karras2019style} & & & 4.4 & \\
    Score SDE~\cite{song2021scorebased} & 65.57M & 266.4G & 7.2\\
    %  LSDM~\cite{vahdat2021score}  & & & 7.2\\
     DDGAN~\cite{vahdat2021score} & 39.73M& 69.9G& 7.6\\
    %  IDDPM & \\
    %  ADM-G & \\
    LDM~\cite{rombach2022high} & 274.1M & 96.1G & 5.1\\
     \hline
    %  GAN-Slim~\cite{wang2020gan} & GAN & & 5.0B & 12.4 \\
    %  CA-Comp~\cite{liu2021content} & GAN & & 4.1B & 7.9 \\
     Lite-LDM & 22.4M({\tiny\textcolor{Green}{$12.2\times $}}) & 7.9G({\tiny\textcolor{Green}{$12.2\times $}}) & 14.3({\tiny\textcolor{red}{$-9.2$}})\\
     Ours & 21.1M({\tiny\textcolor{Green}{$ 13.0\times$}}) & 6.7G({\tiny\textcolor{Green}{$14.3\times $}})& 9.3({\tiny\textcolor{Orange}{$-4.2$}})\\
     \hline
    \end{tabular}}
    {\setlength{\tabcolsep}{1.2pt}
    \footnotesize
        \begin{tabular}{l|l|l|c}
    \hline
    \multicolumn{4}{c}{\textbf{LSUN-Bedroom $256\times 256$}}\\
    \hline
     Model & \#Param & MACs & FID$\downarrow$  \\
      \hline
    %   PGGAN~\cite{karras2018progressive} & GAN & 23.05M & 28.17G & 8.34\\
    %   StyleGAN~\cite{karras2019style} & GAN &25.00M & & 2.35\\
    %  VQGAN~\cite{esser2021taming} & GAN+AR & & & \\
    %  StyleGAN~\cite{karras2019style} & & & 4.4 & \\
    % Score SDE~\cite{song2021scorebased} & Diff & & & \\
    %  LSDM~\cite{vahdat2021score} & Diff & & & \\
    %  DDGAN~\cite{vahdat2021score} & Diff+GAN & & & \\
    DDPM~\cite{ho2020denoising} &  113.7M &248.7G &4.9\\
     IDDPM~\cite{nichol2021improved} &  113.7M &248.6G &4.2\\
     ADM~\cite{dhariwal2021diffusion} & 552.8M & 1114.2G & 1.9\\
    LDM~\cite{rombach2022high} & 274.1M& 96.1G& 3.0\\
     \hline
    %  GAN-Slim~\cite{wang2020gan} & GAN & & 5.0B & 12.4 \\
    %  CA-Comp~\cite{liu2021content} & GAN & & 4.1B & 7.9 \\
     Lite-LDM &22.4M({\tiny\textcolor{Green}{$12.2\times $}}) & 7.9G({\tiny\textcolor{Green}{$12.2\times $}}) & 10.9({\tiny\textcolor{red}{$-7.9$}})\\
     Ours &21.1M({\tiny\textcolor{Green}{$ 13.0\times$}})&6.7G(\textcolor{Green}{{\tiny$14.3\times $}}) &5.2({\tiny\textcolor{Orange}{$-2.2$}})\\
     \hline
    \end{tabular}}
    \hspace{0.3em}
    {\setlength{\tabcolsep}{3pt}
    \footnotesize
\begin{tabular}{l|l|l|c}
    \hline
    \multicolumn{4}{c}{\textbf{LSUN-Church $256\times 256$}}\\
    \hline
     Model & \#Param & MACs & FID$\downarrow$  \\
      \hline
    %   PGGAN~\cite{karras2018progressive} & GAN & 23.05M & 28.17G& 8.34\\
    %   StyleGAN~\cite{karras2019style} & GAN & 25.00M & & 2.35\\
    %  VQGAN~\cite{esser2021taming} & GAN+AR & & & \\
    %  StyleGAN~\cite{karras2019style} & & & 4.4 & \\
    % Score SDE~\cite{song2021scorebased} & Diff & & & \\
    %  LSDM~\cite{vahdat2021score} & Diff & & & \\
    %  DDGAN~\cite{vahdat2021score} & Diff+GAN & & & \\
    DDPM~\cite{ho2020denoising} &  113.7M &248.7G & 4.9\\
     IDDPM~\cite{nichol2021improved}  &113.7M &248.6G &4.3\\
     ADM~\cite{dhariwal2021diffusion} & 552.8M & 1114.2G & 1.9\\
    LDM~\cite{rombach2022high}  & 295.0M & 18.7G &  4.0\\
     \hline
    %  GAN-Slim~\cite{wang2020gan} & GAN & & 5.0B & 12.4 \\
    %  CA-Comp~\cite{liu2021content} & GAN & & 4.1B & 7.9 \\
     Lite-LDM & 32.8M({\tiny\textcolor{Green}{$9.0\times $}}) & 2.1G({\tiny\textcolor{Green}{$8.9\times $}}) & 13.6({\tiny\textcolor{Red}{$-9.6$}})\\
     Ours & 33.8M({\tiny\textcolor{Green}{$8.7\times $}}) & 2.1G({\tiny\textcolor{Green}{$8.9\times $}})&8.4({\tiny\textcolor{Orange}{$-4.4$}})\\
     \hline
    \end{tabular}}
    \caption{Unconditional generation results comparison to prior DPMs. The results are taken from the original paper, except that DDPM is take from the~\cite{choi2022perception}.}
    \label{tab:unconditional}
\end{minipage}
 \hspace{0.1em}
\begin{minipage}{0.26\textwidth}
% \captionsetup[figure]{}
\centering
\vspace{-2.5mm}
\includegraphics[width=\linewidth]{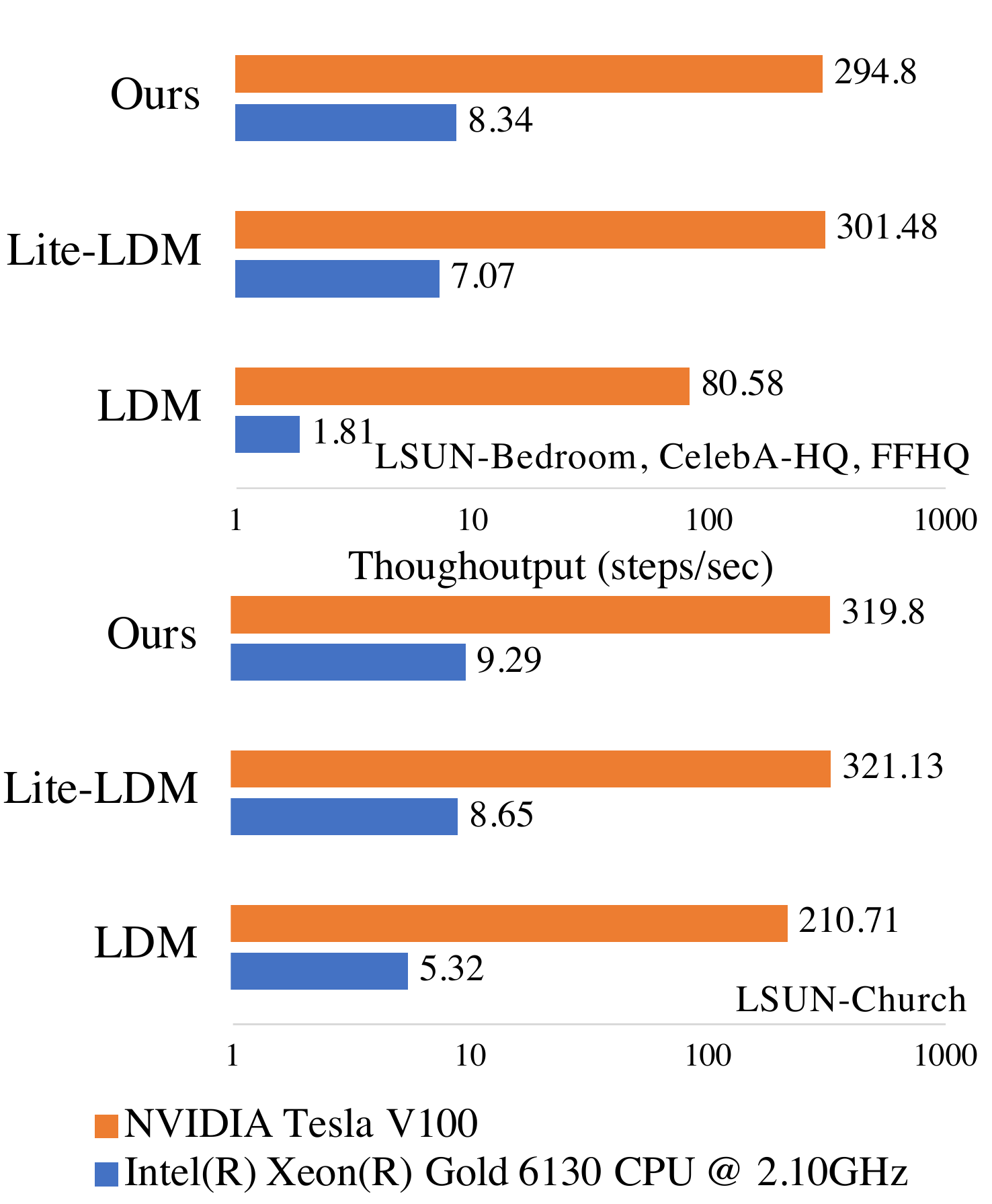}
\captionof{figure}{Throughput for unconditional image generation.}
% \caption{Thoughoutput for unconditional image generation.}
\label{pic:througoutput}
\end{minipage}
\vspace{-6mm}
\end{table*}

\noindent\textbf{Datasets and Evaluation.} We evaluate our model on 4 unconditional generation datasets and 2 conditional benckmarks. Specially, we train our unconditional S\texttt{SD} models on LSUN-Churches/Bedrooms~\cite{yu2015lsun},  FFHQ~\cite{karras2019style}, and CelebA-HQ~\cite{karras2017progressive}. We also validate the model on class-conditioned ImageNet~\cite{deng2009imagenet} and MS-COCO~\cite{lin2014microsoft} text-to-image generation. For the text-to-image task, we first train on LAION-400M~\cite{DBLP:journals/corr/abs-2111-02114} and test on MS-COCO directly. 

\noindent\textbf{Training and Evaluation Details.} We build our model on the LDM~\cite{rombach2022high} frameworks. All pre-trained teachers and auto-encoders are downloaded from the official repository\footnote{https://github.com/CompVis/latent-diffusion}. For fair comparison\footnote{\yxy{Generative models from other families~(e.g. GAN, VAE, and Flow) are excluded intentionally for fair computation comparison.}}, we implement a lite-version of LDM, with a channel dimension of $64$ as our baseline model. We call it Lite-LDM.

On 4 unconditional benchmarks, we train our spectral diffusion for 150k iterations with a mini-batch size of $512$. We use AadmW~\cite{loshchilov2017decoupled} optimizer with initial learning rate \num{1.024e-3} and linear lr decay. For the class- and text-conditioned generation, the initial learning rate is set to \num{5.12e-4} instead, with other parameters unchanged. Classifier-free guidance~\cite{ho2022classifier} is applied. The synthesized image quality is measured by the FID score~\cite{heusel2017gans} with 50k generated samples at the resolution of $256$. We use a 200-step DDIM~\cite{song2020denoising} sampling by default. We also compare the model size and computational cost in terms of parameter number and Multiply-Add cumulation~(MACs)\footnote{https://github.com/sovrasov/flops-counter.pytorch}. Throughput is reported as our measurement of running speed. All experiments are run on 8 NVIDIA Tesla V100 GPUs. More details are specified in the Supplementary Material.

\begin{figure*}
    \centering
    % \includegraphics[width=0.48\linewidth]{figs/celebahq.pdf}
    % \hspace{0.05em}
    % \includegraphics[width=0.48\linewidth]{figs/ffhq.pdf}
    % \includegraphics[width=0.48\linewidth]{figs/bedroom.pdf}\hspace{0.1em}
    % \includegraphics[width=0.48\linewidth]{figs/church.pdf}
    \includegraphics[width=\linewidth]{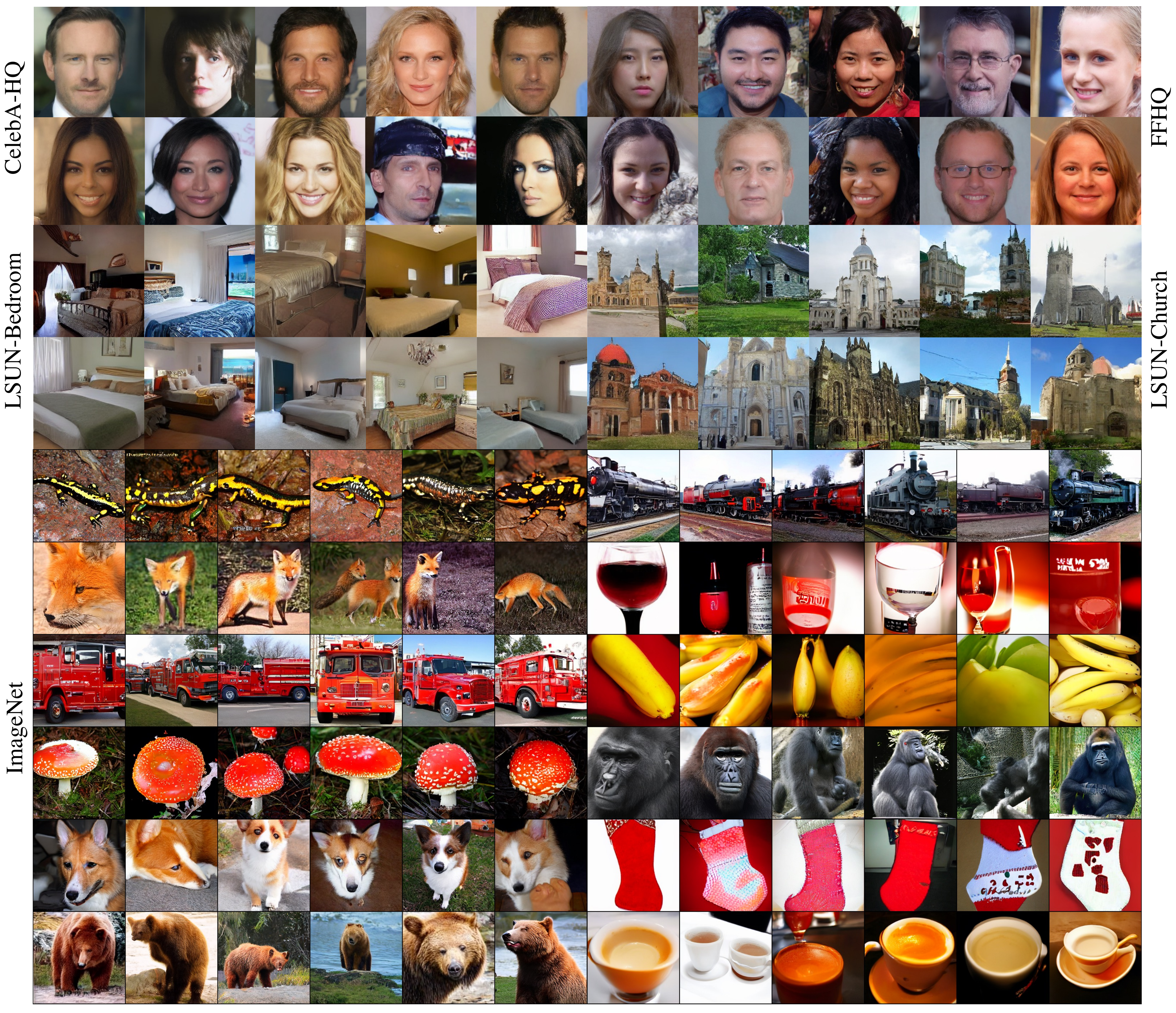}
    \vspace{-7mm}
    \caption{Randomly sampled $256\times 256$ images generated by our models trained on CelebA-HQ~\cite{karras2017progressive}, FFHQ~\cite{karras2019style}, LSUN-Bedroom and LSUN-Church~\cite{yu2015lsun}, ImageNet~\cite{deng2009imagenet}. All images are sampled with 200 DDIM steps.}
    \label{fig:con_uncon_vis}
    \vspace{-6mm}
\end{figure*}
\subsection{Image Generation Results}\label{sec:uncondition}

\textbf{Unconditional Image Generation.}
We train our \texttt{SD} on LSUN-Churches/Bedrooms~\cite{yu2015lsun} FFHQ~\cite{karras2019style}, and CelebA-HQ~\cite{karras2017progressive}, and evaluate the sample quality. As shown in Table~\ref{tab:unconditional}, directly training small-sized diffusion models largely deteriorates the model performance, such that Lite-LDM achieves an FID drop of $12.3$ on FFHQ and $13.2$ on CelebA-HQ. Our proposed \texttt{SD} achieves $8\sim 14$ times parameter and computation reduction compared to official LDM while being competitive in image fidelity. For example, with a 21.1M Unet model and 6.7G MACs, our \texttt{SD} gets an FID score of 5.2, which is very close to the 4.9 FID in DDPM, but with only $\frac{1}{37}$ of its computation cost. 

Throughput is reported in  Figure~\ref{pic:througoutput}. It refers to the number of time steps that model runs per second. We measure its value with a batch-size of 64 by averaging over 30 runs. We see that, Lite-LDM, while being fast, suffer greatly from low visual quality. In comparison, our \texttt{SD} is $4.6\times$ faster on CPU and $3.6\times$ on GPU compared to LDM on 3 of the 4 datasets. 

We inspect the visual quality of the synthesized sample in Figure~\ref{fig:con_uncon_vis}, row 1-4. With much less parameters and complexity, our \texttt{SD} still produces realistic samples with decent high-frequency details and sample diversity.

\textbf{Class-conditional Image Generation.} We validate our performance for class-conditioned image generation on ImageNet. The results are demonstrated in Table~\ref{tab:imagenet}. With super-mini architecture and classifier-free guidance of $w=3.0$, our \texttt{SD} reaches an FID score of 10.6. As the comparison, the ADM~\cite{dhariwal2021diffusion} only gets FID=10.9, but with 553.8M parameters and 1114.2 MACs. Lite-LDM, though being comparably fast, suffers from its inability for high-frequency generation, gets a high FID score of 20.1.

Generated results are visualized in Figure~\ref{fig:con_uncon_vis} row 5-10. Our \texttt{SD} is able to produce diverse images of different categories, particularly good at animal generation like \texttt{corgi} and \texttt{bear}. However, we still observe failure cases with distorted faces and shapes. For example, our models suffer in crowded instance generation such as on \texttt{banana}.\\
\begin{table}[]
\footnotesize
    \centering
    \begin{tabular}{l|l|l|c}
    \hline
        Method & \#Param & MACs & FID$\downarrow$\\
        \hline
        % BigGAN-deep~\cite{brock2018large} & 340M & & 6.95\\
        IDDPM~\cite{nichol2021improved} & 273.1M &  1416.3G & 	12.3\\
        ADM~\cite{dhariwal2021diffusion} & 553.8M & 1114.2G & 10.9 \\
        LDM~\cite{rombach2022high} & 400.9M & 99.8G & 10.6\\
        ADM-G~\cite{dhariwal2021diffusion} & 553.8+54.1M & 1114.2+72.2G & 4.6 \\
        LDM-CFG~\cite{rombach2022high} & 400.9M & 99.8G & 3.6\\\hline
         Lite-LDM-CFG & 47.0M(\textcolor{Green}{$8.5\times$}) & 11.1G (\textcolor{Green}{$9.0\times $})& 20.1(\textcolor{red}{$- 16.5$})\\
        Ours-CFG & 45.4M(\textcolor{Green}{$ 8.8\times$}) & 9.9G (\textcolor{Green}{$10.1\times $})& 10.6(\textcolor{Orange}{$- 7.0$})\\
        \hline
    \end{tabular}
    \vspace{-2mm}
    \caption{Comparison of class-conditional image generation methods on ImageNet~\cite{deng2009imagenet} with recent state-of-the-art methods. ``G'' stands for the classifier guidance and ``CFG'' refers to the classifer-free guidance for conditional image generation.}
    \label{tab:imagenet}
    \vspace{-4mm}
\end{table}

% \begin{figure}
%     \centering
%     \includegraphics[width=\linewidth]{figs/imagenet_models.pdf}
%     \vspace{-4mm}
%     \caption{Comparison of class-conditional image generation methods on ImageNet~\cite{deng2009imagenet} with recent state-of-the-art methods. ``G'' stands for the classifier guidance and ``CFG'' refers to the classifer-free guidance for conditional image generation.}
%     \vspace{-4mm}
%     \label{tab:imagenet}
% \end{figure}[]

\begin{table}[]
\footnotesize
    \centering
    \begin{tabular}{l|l|c}
    \hline
        Method & \#Param &FID$\downarrow$ \\
        \hline
        GLIDE~\cite{nichol2021glide} &  5.0B &  12.24\\
    DALLE2~\cite{ramesh2022hierarchical} & 5.5B & 10.39\\
        Imagen~\cite{saharia2022photorealistic} & 3.0B & 7.27\\
        LDM~\cite{rombach2022high} & 1.45B & 12.63\\
        \hline
        Ours & 77.6M(\textcolor{Green}{$18.7\times$}) & 18.43\\
        \hline
    \end{tabular}
    \vspace{-2mm}
    \caption{Zero-Shot evaluation on MS-COCO text-to-image generation. We only count the model size of diffusion part but exclude language encoder. }
    \label{tab:mscoco_t2i}
    \vspace{-4mm}
\end{table}
\begin{figure*}
    \centering
    \includegraphics[width=\linewidth]{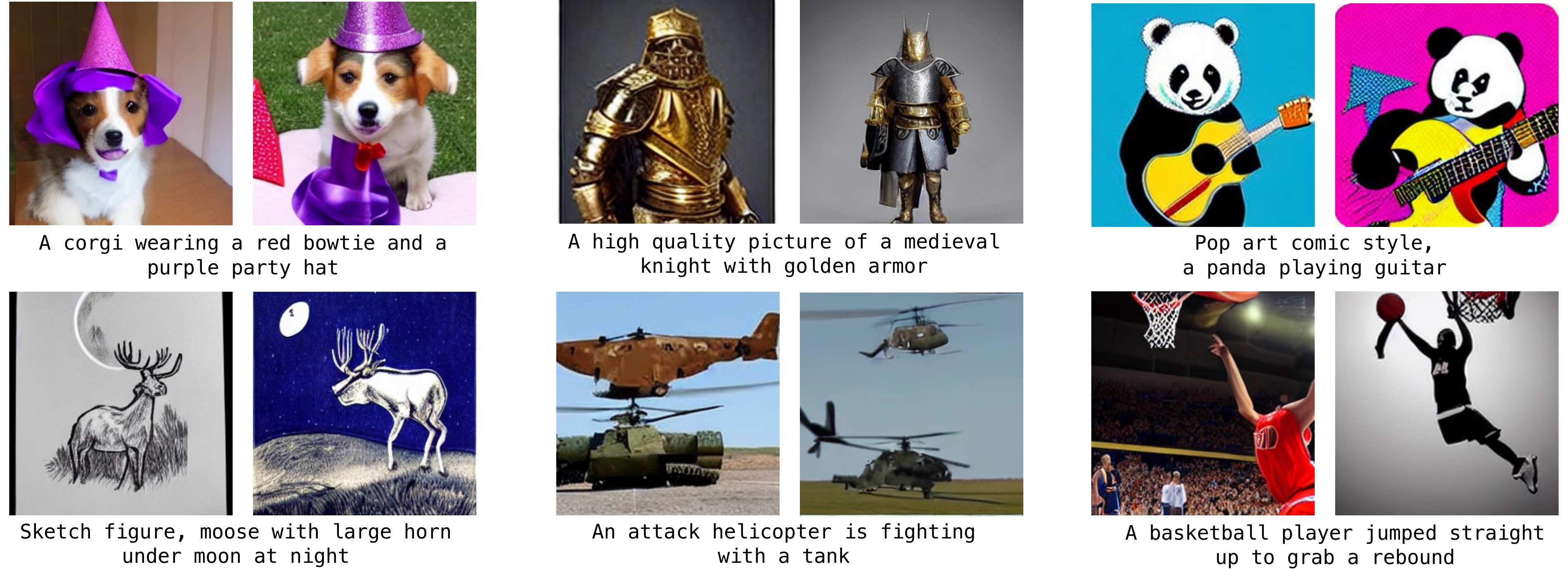}
      \vspace{-6mm}
    \caption{Selected samples from Spectral Diffusion using classifier-free guidance $w=5.0$ for text-to-image generation.}
    \label{fig:t2i}
    \vspace{-4mm}
\end{figure*}
\textbf{Text-to-Image Generation.} Following prior work~\cite{rombach2022high}, we train our text-conditioned \texttt{SD} with a fixed CLIP encoder~\cite{radford2021learning} on LAION-400M~\cite{DBLP:journals/corr/abs-2111-02114}, and then do zero-shot inference on MS-COCO~\cite{lin2014microsoft} with $w=2.0$. Since each MS-COCO images contains multiple captions, during evaluations, we randomly select 50k  descriptions from the train set, with one caption corresponding to a unique image.

The evaluation results are provided in Table~\ref{tab:mscoco_t2i}. Again, with a 77.6M model, we gets to a FID score of 18.43, while being $18.7\times$ smaller than LDM. We also provide qualitative analysis for text-to-image generation with new prompts, in Figure~\ref{fig:t2i}. Although the image quality is not as perfect as in those large-sized diffusion models, our model learns to compose vivid drawing according to the descriptions, with minimal computational cost and portable model size. Our \texttt{SD} is good at abstract or carton style paintings. However, it is still challenging to generate human body and faces, as in the ``basketball player'' example.

\subsection{Ablation Study and Analysis}
\label{sec:ablation}
In this section, we validate the effectiveness of wavelet gating and spectrum-aware distillation, on whether and how they help to improve the image fidelity. 

\noindent\textbf{Effectiveness of Wavelet Gating.} We validate the effectiveness of the Wavelet Gating by replacing our WG upsample and downsample with the nearest neighbor resizer in LDM~\cite{rombach2022high} and train on the FFHQ dataset. As shown in Table~\ref{tab:ablation}, removing WG significantly increases the FID from $10.5\to 12.4$. Besides, WG alone improves Lite-LDM's FID score by $2.6$. Both results indicate that WG 

effectively promote the sample quality of the small DPMs.

\begin{figure}
    \centering
    \includegraphics[width=\linewidth]{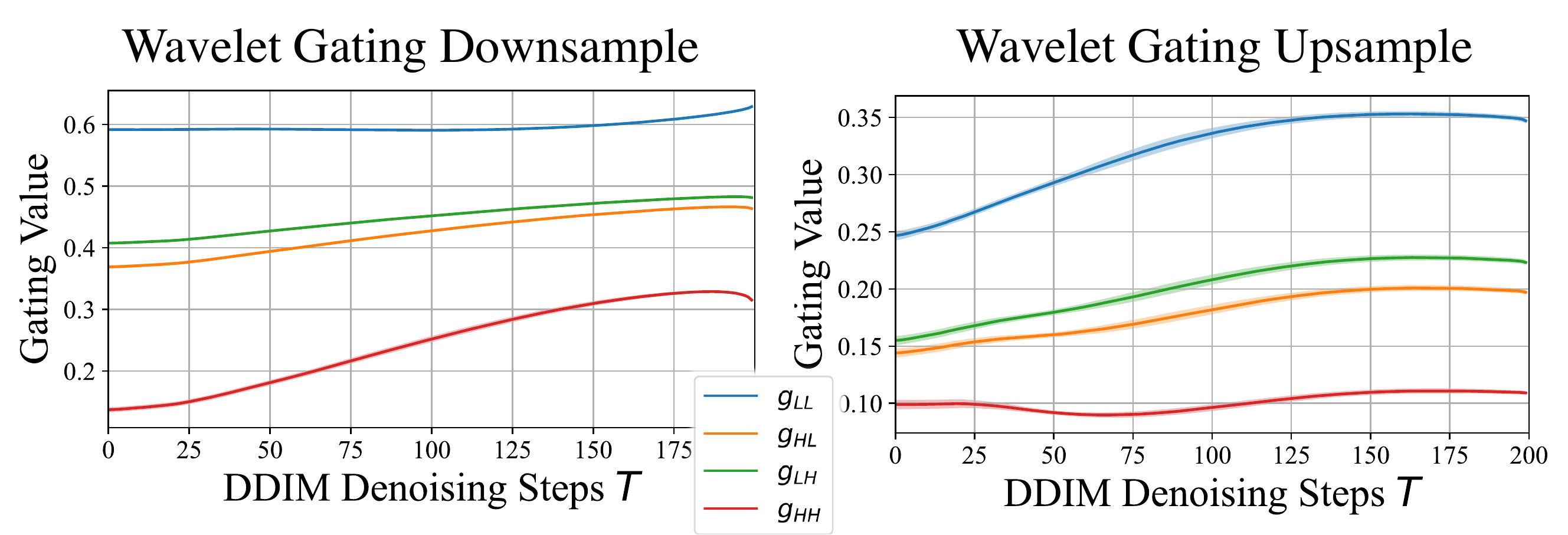}
    \vspace{-7mm}
    \caption{Wavelet gating function values at different $t$. We plot the mean$\pm$std for 100 generated images.}
    \label{fig:wave_gate_dynamics}
     \vspace{-2mm}
\end{figure}

In addition, we plot the values of the gating functions at different denoising steps for a pre-trained text-to-image \texttt{SD} model in Figure~\ref{fig:wave_gate_dynamics}. Each curve is calculated by averaging the gating coefficient for 100 generated images. The trends of the downsample and upsample operations diverge. In the end of denoising~(large $t$), high-frequency details emerged in $\hat{\mathbf{x}}_t$. The \texttt{WG-Down} thus enhances the high-frequency signals with increased $g_{\{\textsf{HL},\textsf{LH},\textsf{HH}\}}$ while keeping the low-frequency part constant. In contrast, the \texttt{WG-Up}~(Right) promotes $g_{\textsf{LL}}$ in the late stage of denoising. Predicted noises boost its low-frequency components, resulting in high-frequency element recovery in the $\hat{\mathbf{x}}_0=\frac{\mathbf{x}_t-\sqrt{1-\Bar{\alpha} \bm \epsilon}}{\sqrt{\Bar{\alpha}}}$.

% It reflects the dynamic frequency response of the denoising network at different time-step.
\noindent\textbf{Effectiveness of Spectrum-Aware Distillation.} To understand the value of the proposed SA-Distillation, we sequentially remove each loss term. Figure~\ref{tab:ablation} shows that, while the spatial term only accounts for 0.9 FID, the frequency term takes up 1.8 FID improvement, highlighting its importance in high-quality image generation.

\begin{figure}
    \centering
    \includegraphics[width=\linewidth]{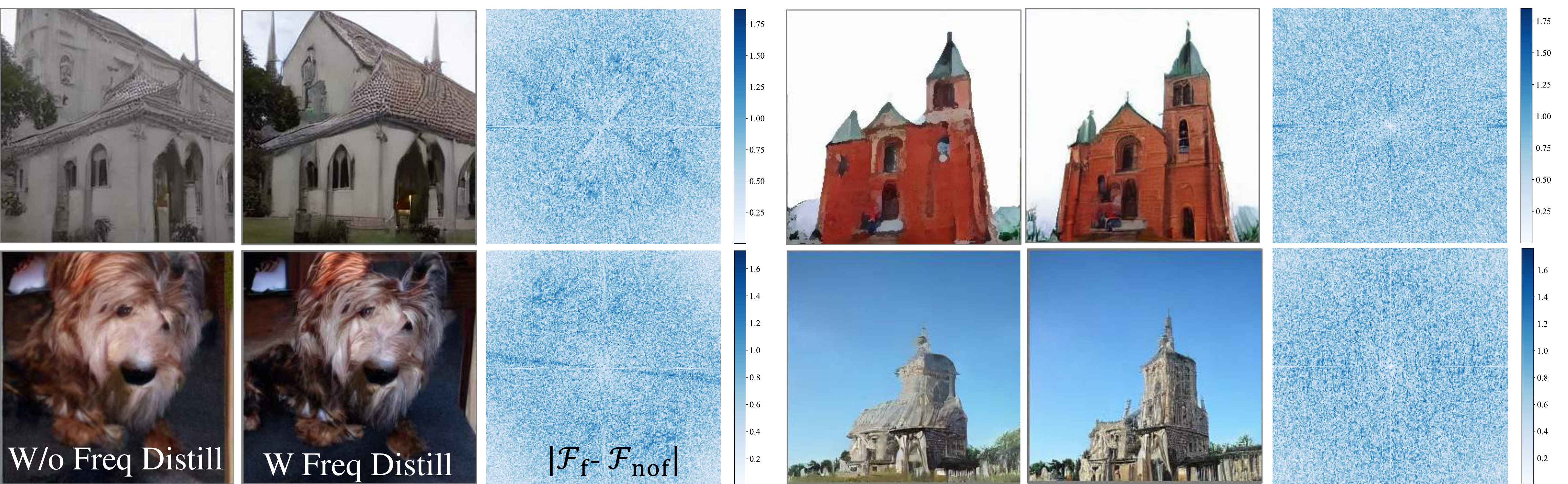}
    \vspace{-5mm}
    \caption{Generated images W or W/O the freq term, as well as their DFT difference $|\mathcal{F}_{\text{f}}-\mathcal{F}_{\text{nof}}|$.}
    \label{fig:distill_ab}
    \vspace{-5mm}
\end{figure}

We also visualize the images generated by trained models with~(W) or without~(W/O) the frequency term in Figure~\ref{fig:distill_ab}, with their DFT difference. The model without $\mathcal{L}_{freq}$ makes smoother predictions, while our method recovers the details like hair or architectural textures. By penalizing high-frequency distillation, our proposed SA-Distillation resulted in large differences and improvements in high-frequency components in $|\mathcal{F}_{f}-\mathcal{F}_{nof}|$.

\begin{table}[]
\setlength{\tabcolsep}{0.2em}
\renewcommand{\arraystretch}{1.}
\footnotesize
    \centering
    \begin{tabular}{l|c c c c c c c c}
    \hline
        Method & \multicolumn{7}{c}{\textbf{FFHQ $256\times 256$}} \\
        \hline
       + Wavelet Gating  &   & \textcolor{Green}{\Checkmark} &  & & \textcolor{Green}{\Checkmark} & & \textcolor{Green}{\Checkmark} & \textcolor{Green}{\Checkmark}\\
       + Spatial Distill &  &  & \textcolor{Green}{\Checkmark} &  & \textcolor{Green}{\Checkmark} & \textcolor{Green}{\Checkmark} &  &\textcolor{Green}{\Checkmark}\\
       + Freq Distill & & &  & \textcolor{Green}{\Checkmark} &  &\textcolor{Green}{\Checkmark}& \textcolor{Green}{\Checkmark} & \textcolor{Green}{\Checkmark}\\
       \hline
       FID$\downarrow$ & 17.3 & 14.7 & 16.6 & 15.3 & 12.3 &  12.4& 11.4&10.5\\\hline
    \end{tabular}
    \vspace{-3mm}
    \caption{Ablation study on FFHQ dataset.}
    \label{tab:ablation}
    \vspace{-3mm}
\end{table}

\section{Conclusion}
In the study, we focus on reducing the computation cost for diffusion models. The primary obstacle to training small DPMs is their inability to provide realistic high-frequency, which results from the frequency evolution and bias of diffusion process. In order to resolve these problems, we propose Spectral Diffusion~(\texttt{SD}) for efficient image generation. It performs spectrum dynamic denoising by using a wavelet gating operation, which automatically enhances different frequency bands at different reverse steps. A large pre-trained network helps to improve the performance of high-frequency generation by knowledge distillation. By seamlessly integrating both modifications, our model is 8-18 $\times$ slimer and runs 2-5$\times$ faster than the latent diffusion model, with negligible performance drop. 
\newpage
% \textbf{Progressive Training}
%%%%%%%%% REFERENCES
{\small
\bibliographystyle{ieee_fullname}
\bibliography{egbib}

\begin{thebibliography}{10}\itemsep=-1pt

\bibitem{bao2022analytic}
Fan Bao, Chongxuan Li, Jun Zhu, and Bo Zhang.
\newblock Analytic-dpm: an analytic estimate of the optimal reverse variance in
  diffusion probabilistic models.
\newblock {\em arXiv preprint arXiv:2201.06503}, 2022.

\bibitem{basri2020frequency}
Ronen Basri, Meirav Galun, Amnon Geifman, David Jacobs, Yoni Kasten, and Shira
  Kritchman.
\newblock Frequency bias in neural networks for input of non-uniform density.
\newblock In {\em International Conference on Machine Learning}, pages
  685--694. PMLR, 2020.

\bibitem{burton1987color}
Geoffrey~J Burton and Ian~R Moorhead.
\newblock Color and spatial structure in natural scenes.
\newblock {\em Applied optics}, 26(1):157--170, 1987.

\bibitem{cao2019learning}
Kaidi Cao, Colin Wei, Adrien Gaidon, Nikos Arechiga, and Tengyu Ma.
\newblock Learning imbalanced datasets with label-distribution-aware margin
  loss.
\newblock {\em Advances in neural information processing systems}, 32, 2019.

\bibitem{chen2021ssd}
Yuanqi Chen, Ge Li, Cece Jin, Shan Liu, and Thomas Li.
\newblock Ssd-gan: Measuring the realness in the spatial and spectral domains.
\newblock In {\em Proceedings of the AAAI Conference on Artificial
  Intelligence}, volume~35, pages 1105--1112, 2021.

\bibitem{choi2022perception}
Jooyoung Choi, Jungbeom Lee, Chaehun Shin, Sungwon Kim, Hyunwoo Kim, and
  Sungroh Yoon.
\newblock Perception prioritized training of diffusion models.
\newblock In {\em Proceedings of the IEEE/CVF Conference on Computer Vision and
  Pattern Recognition}, pages 11472--11481, 2022.

\bibitem{deng2009imagenet}
Jia Deng, Wei Dong, Richard Socher, Li-Jia Li, Kai Li, and Li Fei-Fei.
\newblock Imagenet: A large-scale hierarchical image database.
\newblock In {\em 2009 IEEE conference on computer vision and pattern
  recognition}, pages 248--255. Ieee, 2009.

\bibitem{dhariwal2021diffusion}
Prafulla Dhariwal and Alexander Nichol.
\newblock Diffusion models beat gans on image synthesis.
\newblock {\em Advances in Neural Information Processing Systems},
  34:8780--8794, 2021.

\bibitem{field1987relations}
David~J Field.
\newblock Relations between the statistics of natural images and the response
  properties of cortical cells.
\newblock {\em Josa a}, 4(12):2379--2394, 1987.

\bibitem{frank2020leveraging}
Joel Frank, Thorsten Eisenhofer, Lea Sch{\"o}nherr, Asja Fischer, Dorothea
  Kolossa, and Thorsten Holz.
\newblock Leveraging frequency analysis for deep fake image recognition.
\newblock In {\em International conference on machine learning}, pages
  3247--3258. PMLR, 2020.

\bibitem{fu2021dw}
Minghan Fu, Huan Liu, Yankun Yu, Jun Chen, and Keyan Wang.
\newblock Dw-gan: A discrete wavelet transform gan for nonhomogeneous dehazing.
\newblock In {\em Proceedings of the IEEE/CVF Conference on Computer Vision and
  Pattern Recognition}, pages 203--212, 2021.

\bibitem{goodfellow2020generative}
Ian Goodfellow, Jean Pouget-Abadie, Mehdi Mirza, Bing Xu, David Warde-Farley,
  Sherjil Ozair, Aaron Courville, and Yoshua Bengio.
\newblock Generative adversarial networks.
\newblock {\em Communications of the ACM}, 63(11):139--144, 2020.

\bibitem{gu2022vector}
Shuyang Gu, Dong Chen, Jianmin Bao, Fang Wen, Bo Zhang, Dongdong Chen, Lu Yuan,
  and Baining Guo.
\newblock Vector quantized diffusion model for text-to-image synthesis.
\newblock In {\em Proceedings of the IEEE/CVF Conference on Computer Vision and
  Pattern Recognition}, pages 10696--10706, 2022.

\bibitem{heusel2017gans}
Martin Heusel, Hubert Ramsauer, Thomas Unterthiner, Bernhard Nessler, and Sepp
  Hochreiter.
\newblock Gans trained by a two time-scale update rule converge to a local nash
  equilibrium.
\newblock {\em Advances in neural information processing systems}, 30, 2017.

\bibitem{ho2022imagen}
Jonathan Ho, William Chan, Chitwan Saharia, Jay Whang, Ruiqi Gao, Alexey
  Gritsenko, Diederik~P Kingma, Ben Poole, Mohammad Norouzi, David~J Fleet,
  et~al.
\newblock Imagen video: High definition video generation with diffusion models.
\newblock {\em arXiv preprint arXiv:2210.02303}, 2022.

\bibitem{ho2020denoising}
Jonathan Ho, Ajay Jain, and Pieter Abbeel.
\newblock Denoising diffusion probabilistic models.
\newblock {\em Advances in Neural Information Processing Systems},
  33:6840--6851, 2020.

\bibitem{ho2022cascaded}
Jonathan Ho, Chitwan Saharia, William Chan, David~J Fleet, Mohammad Norouzi,
  and Tim Salimans.
\newblock Cascaded diffusion models for high fidelity image generation.
\newblock {\em J. Mach. Learn. Res.}, 23:47--1, 2022.

\bibitem{ho2022classifier}
Jonathan Ho and Tim Salimans.
\newblock Classifier-free diffusion guidance.
\newblock {\em arXiv preprint arXiv:2207.12598}, 2022.

\bibitem{ho2022video}
Jonathan Ho, Tim Salimans, Alexey Gritsenko, William Chan, Mohammad Norouzi,
  and David~J Fleet.
\newblock Video diffusion models.
\newblock {\em arXiv:2204.03458}, 2022.

\bibitem{hu2018squeeze}
Jie Hu, Li Shen, and Gang Sun.
\newblock Squeeze-and-excitation networks.
\newblock In {\em Proceedings of the IEEE conference on computer vision and
  pattern recognition}, pages 7132--7141, 2018.

\bibitem{huang2019ccnet}
Zilong Huang, Xinggang Wang, Lichao Huang, Chang Huang, Yunchao Wei, and Wenyu
  Liu.
\newblock Ccnet: Criss-cross attention for semantic segmentation.
\newblock In {\em Proceedings of the IEEE/CVF international conference on
  computer vision}, pages 603--612, 2019.

\bibitem{hyvarinen2005estimation}
Aapo Hyv{\"a}rinen and Peter Dayan.
\newblock Estimation of non-normalized statistical models by score matching.
\newblock {\em Journal of Machine Learning Research}, 6(4), 2005.

\bibitem{jiang2021focal}
Liming Jiang, Bo Dai, Wayne Wu, and Chen~Change Loy.
\newblock Focal frequency loss for image reconstruction and synthesis.
\newblock In {\em Proceedings of the IEEE/CVF International Conference on
  Computer Vision}, pages 13919--13929, 2021.

\bibitem{Kang2020Decoupling}
Bingyi Kang, Saining Xie, Marcus Rohrbach, Zhicheng Yan, Albert Gordo, Jiashi
  Feng, and Yannis Kalantidis.
\newblock Decoupling representation and classifier for long-tailed recognition.
\newblock In {\em International Conference on Learning Representations}, 2020.

\bibitem{karras2017progressive}
Tero Karras, Timo Aila, Samuli Laine, and Jaakko Lehtinen.
\newblock Progressive growing of gans for improved quality, stability, and
  variation.
\newblock {\em arXiv preprint arXiv:1710.10196}, 2017.

\bibitem{karras2019style}
Tero Karras, Samuli Laine, and Timo Aila.
\newblock A style-based generator architecture for generative adversarial
  networks.
\newblock In {\em Proceedings of the IEEE/CVF conference on computer vision and
  pattern recognition}, pages 4401--4410, 2019.

\bibitem{khayatkhoei2022spatial}
Mahyar Khayatkhoei and Ahmed Elgammal.
\newblock Spatial frequency bias in convolutional generative adversarial
  networks.
\newblock In {\em Proceedings of the AAAI Conference on Artificial
  Intelligence}, volume~36, pages 7152--7159, 2022.

\bibitem{lin2017focal}
Tsung-Yi Lin, Priya Goyal, Ross Girshick, Kaiming He, and Piotr Doll{\'a}r.
\newblock Focal loss for dense object detection.
\newblock In {\em Proceedings of the IEEE international conference on computer
  vision}, pages 2980--2988, 2017.

\bibitem{lin2014microsoft}
Tsung-Yi Lin, Michael Maire, Serge Belongie, James Hays, Pietro Perona, Deva
  Ramanan, Piotr Doll{\'a}r, and C~Lawrence Zitnick.
\newblock Microsoft coco: Common objects in context.
\newblock In {\em European conference on computer vision}, pages 740--755.
  Springer, 2014.

\bibitem{liu2022pseudo}
Luping Liu, Yi Ren, Zhijie Lin, and Zhou Zhao.
\newblock Pseudo numerical methods for diffusion models on manifolds.
\newblock In {\em International Conference on Learning Representations}, 2022.

\bibitem{liu2019structured}
Yifan Liu, Ke Chen, Chris Liu, Zengchang Qin, Zhenbo Luo, and Jingdong Wang.
\newblock Structured knowledge distillation for semantic segmentation.
\newblock In {\em Proceedings of the IEEE/CVF Conference on Computer Vision and
  Pattern Recognition}, pages 2604--2613, 2019.

\bibitem{loshchilov2017decoupled}
Ilya Loshchilov and Frank Hutter.
\newblock Decoupled weight decay regularization.
\newblock {\em arXiv preprint arXiv:1711.05101}, 2017.

\bibitem{lu2022dpm}
Cheng Lu, Yuhao Zhou, Fan Bao, Jianfei Chen, Chongxuan Li, and Jun Zhu.
\newblock Dpm-solver: A fast ode solver for diffusion probabilistic model
  sampling in around 10 steps.
\newblock {\em arXiv preprint arXiv:2206.00927}, 2022.

\bibitem{luhman2021knowledge}
Eric Luhman and Troy Luhman.
\newblock Knowledge distillation in iterative generative models for improved
  sampling speed.
\newblock {\em arXiv preprint arXiv:2101.02388}, 2021.

\bibitem{ma2022pds}
Hengyuan Ma, Li Zhang, Xiatian Zhu, and Jianfeng Feng.
\newblock Accelerating score-based generative models with preconditioned
  diffusion sampling.
\newblock In {\em European Conference on Computer Vision}, 2022.

\bibitem{meng2022distillation}
Chenlin Meng, Ruiqi Gao, Diederik~P Kingma, Stefano Ermon, Jonathan Ho, and Tim
  Salimans.
\newblock On distillation of guided diffusion models.
\newblock {\em arXiv preprint arXiv:2210.03142}, 2022.

\bibitem{nichol2021glide}
Alex Nichol, Prafulla Dhariwal, Aditya Ramesh, Pranav Shyam, Pamela Mishkin,
  Bob McGrew, Ilya Sutskever, and Mark Chen.
\newblock Glide: Towards photorealistic image generation and editing with
  text-guided diffusion models.
\newblock {\em arXiv preprint arXiv:2112.10741}, 2021.

\bibitem{nichol2021improved}
Alexander~Quinn Nichol and Prafulla Dhariwal.
\newblock Improved denoising diffusion probabilistic models.
\newblock In {\em International Conference on Machine Learning}, pages
  8162--8171. PMLR, 2021.

\bibitem{poole2022dreamfusion}
Ben Poole, Ajay Jain, Jonathan~T. Barron, and Ben Mildenhall.
\newblock Dreamfusion: Text-to-3d using 2d diffusion.
\newblock {\em arXiv}, 2022.

\bibitem{qin2021fcanet}
Zequn Qin, Pengyi Zhang, Fei Wu, and Xi Li.
\newblock Fcanet: Frequency channel attention networks.
\newblock In {\em Proceedings of the IEEE/CVF international conference on
  computer vision}, pages 783--792, 2021.

\bibitem{radford2021learning}
Alec Radford, Jong~Wook Kim, Chris Hallacy, Aditya Ramesh, Gabriel Goh,
  Sandhini Agarwal, Girish Sastry, Amanda Askell, Pamela Mishkin, Jack Clark,
  et~al.
\newblock Learning transferable visual models from natural language
  supervision.
\newblock In {\em International Conference on Machine Learning}, pages
  8748--8763. PMLR, 2021.

\bibitem{ramesh2022hierarchical}
Aditya Ramesh, Prafulla Dhariwal, Alex Nichol, Casey Chu, and Mark Chen.
\newblock Hierarchical text-conditional image generation with clip latents.
\newblock {\em arXiv preprint arXiv:2204.06125}, 2022.

\bibitem{rombach2022high}
Robin Rombach, Andreas Blattmann, Dominik Lorenz, Patrick Esser, and Bj{\"o}rn
  Ommer.
\newblock High-resolution image synthesis with latent diffusion models.
\newblock In {\em Proceedings of the IEEE/CVF Conference on Computer Vision and
  Pattern Recognition}, pages 10684--10695, 2022.

\bibitem{romero2014fitnets}
Adriana Romero, Nicolas Ballas, Samira~Ebrahimi Kahou, Antoine Chassang, Carlo
  Gatta, and Yoshua Bengio.
\newblock Fitnets: Hints for thin deep nets.
\newblock {\em arXiv preprint arXiv:1412.6550}, 2014.

\bibitem{ronneberger2015u}
Olaf Ronneberger, Philipp Fischer, and Thomas Brox.
\newblock U-net: Convolutional networks for biomedical image segmentation.
\newblock In {\em International Conference on Medical image computing and
  computer-assisted intervention}, pages 234--241. Springer, 2015.

\bibitem{saharia2022photorealistic}
Chitwan Saharia, William Chan, Saurabh Saxena, Lala Li, Jay Whang, Emily
  Denton, Seyed Kamyar~Seyed Ghasemipour, Burcu~Karagol Ayan, S~Sara Mahdavi,
  Rapha~Gontijo Lopes, et~al.
\newblock Photorealistic text-to-image diffusion models with deep language
  understanding.
\newblock {\em arXiv preprint arXiv:2205.11487}, 2022.

\bibitem{salimans2022progressive}
Tim Salimans and Jonathan Ho.
\newblock Progressive distillation for fast sampling of diffusion models.
\newblock In {\em International Conference on Learning Representations}, 2022.

\bibitem{DBLP:journals/corr/abs-2111-02114}
Christoph Schuhmann, Richard Vencu, Romain Beaumont, Robert Kaczmarczyk,
  Clayton Mullis, Aarush Katta, Theo Coombes, Jenia Jitsev, and Aran
  Komatsuzaki.
\newblock {LAION-400M:} open dataset of clip-filtered 400 million image-text
  pairs.
\newblock {\em CoRR}, abs/2111.02114, 2021.

\bibitem{schwarz2021frequency}
Katja Schwarz, Yiyi Liao, and Andreas Geiger.
\newblock On the frequency bias of generative models.
\newblock {\em Advances in Neural Information Processing Systems},
  34:18126--18136, 2021.

\bibitem{sohl2015deep}
Jascha Sohl-Dickstein, Eric Weiss, Niru Maheswaranathan, and Surya Ganguli.
\newblock Deep unsupervised learning using nonequilibrium thermodynamics.
\newblock In {\em International Conference on Machine Learning}, pages
  2256--2265. PMLR, 2015.

\bibitem{song2020denoising}
Jiaming Song, Chenlin Meng, and Stefano Ermon.
\newblock Denoising diffusion implicit models.
\newblock {\em arXiv preprint arXiv:2010.02502}, 2020.

\bibitem{song2021maximum}
Yang Song, Conor Durkan, Iain Murray, and Stefano Ermon.
\newblock Maximum likelihood training of score-based diffusion models.
\newblock {\em Advances in Neural Information Processing Systems},
  34:1415--1428, 2021.

\bibitem{song2019generative}
Yang Song and Stefano Ermon.
\newblock Generative modeling by estimating gradients of the data distribution.
\newblock {\em Advances in Neural Information Processing Systems}, 32, 2019.

\bibitem{song2020sliced}
Yang Song, Sahaj Garg, Jiaxin Shi, and Stefano Ermon.
\newblock Sliced score matching: A scalable approach to density and score
  estimation.
\newblock In {\em Uncertainty in Artificial Intelligence}, pages 574--584.
  PMLR, 2020.

\bibitem{song2021scorebased}
Yang Song, Jascha Sohl-Dickstein, Diederik~P Kingma, Abhishek Kumar, Stefano
  Ermon, and Ben Poole.
\newblock Score-based generative modeling through stochastic differential
  equations.
\newblock In {\em International Conference on Learning Representations}, 2021.

\bibitem{tolhurst1992amplitude}
David~J Tolhurst, Yoav Tadmor, and Tang Chao.
\newblock Amplitude spectra of natural images.
\newblock {\em Ophthalmic and Physiological Optics}, 12(2):229--232, 1992.

\bibitem{vahdat2021score}
Arash Vahdat, Karsten Kreis, and Jan Kautz.
\newblock Score-based generative modeling in latent space.
\newblock {\em Advances in Neural Information Processing Systems},
  34:11287--11302, 2021.

\bibitem{van1996modelling}
van~A Van~der Schaaf and JH~van van Hateren.
\newblock Modelling the power spectra of natural images: statistics and
  information.
\newblock {\em Vision research}, 36(17):2759--2770, 1996.

\bibitem{vaswani2017attention}
Ashish Vaswani, Noam Shazeer, Niki Parmar, Jakob Uszkoreit, Llion Jones,
  Aidan~N Gomez, {\L}ukasz Kaiser, and Illia Polosukhin.
\newblock Attention is all you need.
\newblock {\em Advances in neural information processing systems}, 30, 2017.

\bibitem{vincent2011connection}
Pascal Vincent.
\newblock A connection between score matching and denoising autoencoders.
\newblock {\em Neural computation}, 23(7):1661--1674, 2011.

\bibitem{wiener1949extrapolation}
Norbert Wiener, Norbert Wiener, Cyberneticist Mathematician, Norbert Wiener,
  Norbert Wiener, and Cybern{\'e}ticien Math{\'e}maticien.
\newblock {\em Extrapolation, interpolation, and smoothing of stationary time
  series: with engineering applications}, volume 113.
\newblock MIT press Cambridge, MA, 1949.

\bibitem{woo2018cbam}
Sanghyun Woo, Jongchan Park, Joon-Young Lee, and In~So Kweon.
\newblock Cbam: Convolutional block attention module.
\newblock In {\em Proceedings of the European conference on computer vision
  (ECCV)}, pages 3--19, 2018.

\bibitem{xu2019frequency}
Zhi-Qin~John Xu, Yaoyu Zhang, Tao Luo, Yanyang Xiao, and Zheng Ma.
\newblock Frequency principle: Fourier analysis sheds light on deep neural
  networks.
\newblock {\em arXiv preprint arXiv:1901.06523}, 2019.

\bibitem{xu2019training}
Zhi-Qin~John Xu, Yaoyu Zhang, and Yanyang Xiao.
\newblock Training behavior of deep neural network in frequency domain.
\newblock In {\em International Conference on Neural Information Processing},
  pages 264--274. Springer, 2019.

\bibitem{yang2022wavegan}
Mengping Yang, Zhe Wang, Ziqiu Chi, and Wenyi Feng.
\newblock Wavegan: Frequency-aware gan for high-fidelity few-shot image
  generation.
\newblock {\em arXiv preprint arXiv:2207.07288}, 2022.

\bibitem{yu2015lsun}
Fisher Yu, Ari Seff, Yinda Zhang, Shuran Song, Thomas Funkhouser, and Jianxiong
  Xiao.
\newblock Lsun: Construction of a large-scale image dataset using deep learning
  with humans in the loop.
\newblock {\em arXiv preprint arXiv:1506.03365}, 2015.

\bibitem{zhang2021deep}
Yifan Zhang, Bingyi Kang, Bryan Hooi, Shuicheng Yan, and Jiashi Feng.
\newblock Deep long-tailed learning: A survey.
\newblock {\em arXiv preprint arXiv:2110.04596}, 2021.

\bibitem{zhou2022magicvideo}
Daquan Zhou, Weimin Wang, Hanshu Yan, Weiwei Lv, Yizhe Zhu, and Jiashi Feng.
\newblock Magicvideo: Efficient video generation with latent diffusion models.
\newblock {\em arXiv preprint arXiv:2211.11018}, 2022.

\end{thebibliography}
}

\end{document}